%% file: iclr2027_conference.tex
\documentclass{article} %
\usepackage{iclr2027_conference,times}

\input{math_commands.tex}

\usepackage[utf8]{inputenc} %
\usepackage[T1]{fontenc}    %
\usepackage{hyperref}       %
\usepackage{url}            %
\usepackage{booktabs}       %
\usepackage{makecell}
\usepackage[most]{tcolorbox}
\usepackage{listings}
\usepackage{amsmath}
\usepackage{graphicx}
\usepackage{caption}
\usepackage{subcaption}
\usepackage{multirow}
\usepackage{wrapfig}
\usepackage[table]{xcolor}
\usepackage{tabularx}
\usepackage{tablefootnote}
\usepackage{array}
\usepackage{amsfonts}       %
\usepackage{nicefrac}       %
\usepackage{microtype}      %
\usepackage{xcolor}         %
\usepackage{float}
\usepackage{siunitx}
\usepackage{enumitem}
\usepackage{amssymb}
\usepackage[ruled,vlined]{algorithm2e}
\definecolor{AlgCommentBlue}{RGB}{64,105,150}
\DeclareCaptionFont{subfigfont}{\fontsize{8pt}{10pt}\selectfont}
\SetKwInput{KwInput}{Input}
\SetKwInput{KwOutput}{Output}

\SetCommentSty{AlgCommentSty}
\SetKwComment{Comment}{\textcolor{AlgCommentBlue}{// }}{}

\DontPrintSemicolon
\usepackage{comment}

\definecolor{promptbg}{RGB}{248,249,250}
\definecolor{promptframe}{RGB}{180,180,180}
\definecolor{prompttitle}{RGB}{60,60,60}

\lstdefinestyle{promptstyle}{
    basicstyle=\ttfamily\footnotesize,
    backgroundcolor=\color{promptbg},
    frame=none,
    breaklines=true,
    breakatwhitespace=false,
    columns=fullflexible,
    keepspaces=true,
    showstringspaces=false,
    xleftmargin=0.5em,
    xrightmargin=0.5em
}

\newtcblisting{promptbox}[2][]{
    enhanced,
    breakable,
    colback=promptbg,
    colframe=promptframe,
    boxrule=0.5pt,
    arc=2mm,
    left=1mm,
    right=1mm,
    top=1mm,
    bottom=1mm,
    title={#2},
    coltitle=prompttitle,
    fonttitle=\bfseries,
    listing only,
    listing options={style=promptstyle},
    #1
}

\definecolor{darkblue}{RGB}{0, 0, 123}
\definecolor{bestrow}{RGB}{230, 245, 230}
\usepackage{hyperref}
\hypersetup{
  colorlinks=true,
  urlcolor=darkblue,
  linkcolor=darkblue,
  citecolor=darkblue,
  pdfborder={0 0 0},
  breaklinks=true
}

\title{Beyond Memory Construction: Rethinking Memory Access for LLM-based Conversational Agents}

\author{
\rule{0pt}{2mm}\\[-5mm]
\begin{tabular}{@{}l@{}}
\bfseries
Donghua Cai$^{1}$ \quad
Yongheng Deng$^{1}$\textsuperscript{\textdagger} \quad
Yifei Wang$^{2}$ \quad
Zijun Shen$^{3}$ \quad
Ju Ren$^{1}$
\\[1.5mm]
\normalfont
$^{1}$Tsinghua University, \quad
$^{2}$Beijing Institute of Technology, \quad
$^{3}$Nanjing University
\end{tabular}
}

\iclrfinalcopy
\begin{document}

\maketitle
\lhead{Under review}
\renewcommand{\thefootnote}{}\footnotetext{\textsuperscript{\textdagger}Corresponding Author.}\renewcommand{\thefootnote}{\arabic{footnote}}

\begin{abstract}
Memory is a core component of conversational agents, enabling coherent and context-aware behavior over long interactions. 
Recent approaches commonly rely on LLM-based memory construction, where raw interactions are rewritten into structured memory units and later retrieved via a RAG pipeline. 
While effective in controlled settings, we show that this paradigm breaks down in \emph{long-horizon, high-entropy} conversations: memory construction becomes increasingly lossy and unstable as context length and information complexity grow, and incurs prohibitive cost due to repeated LLM invocation.
To address these limitations, we propose \texttt{Threader}, a memory system that shifts the focus from memory construction to \emph{efficient, structure-aware access over raw interactions}. 
Instead of rewriting interactions, \texttt{Threader} preserves them as first-class memory, organizes them into topic-coherent segments via lightweight incremental segmentation, and enables accurate retrieval through multi-view representation. 
At query time, it performs multi-signal retrieval that combines segment-level access with localized evidence matching, ensuring both completeness and coherence.
Extensive experiments demonstrate that \texttt{Threader} consistently improves answer accuracy and evidence recall, while significantly reducing the memory construction overhead. Code will be available at: \url{https://github.com/Donghua-Cai/Threader}
\end{abstract}

\section{Introduction}
\label{sec:intro}

Large language model (LLM) agents have shown strong capabilities in multi-turn interactions, where maintaining and utilizing long-term memory is essential for coherent, personalized, and context-aware behavior. To support such capabilities, recent works \citep{zhang2025survey,hu2025memory} commonly adopt an external memory paradigm: raw interactions are first transformed into structured memory units via LLM-based rewriting, summarization, or extraction, stored in a memory repository, and later retrieved through a retrieval-augmented generation (RAG) pipeline when answering new queries \citep{chhikara2025mem0,rasmussen2025zep,xu2025mem,hu2026evermemos}. This design offers strong engineering advantages—it avoids modifying backbone models, generalizes across tasks, and can be readily deployed in practical systems.

\begin{figure*}[!tb]
\centering

\begin{subfigure}[t]{0.32\textwidth}
    \centering
    \includegraphics[width=\linewidth]{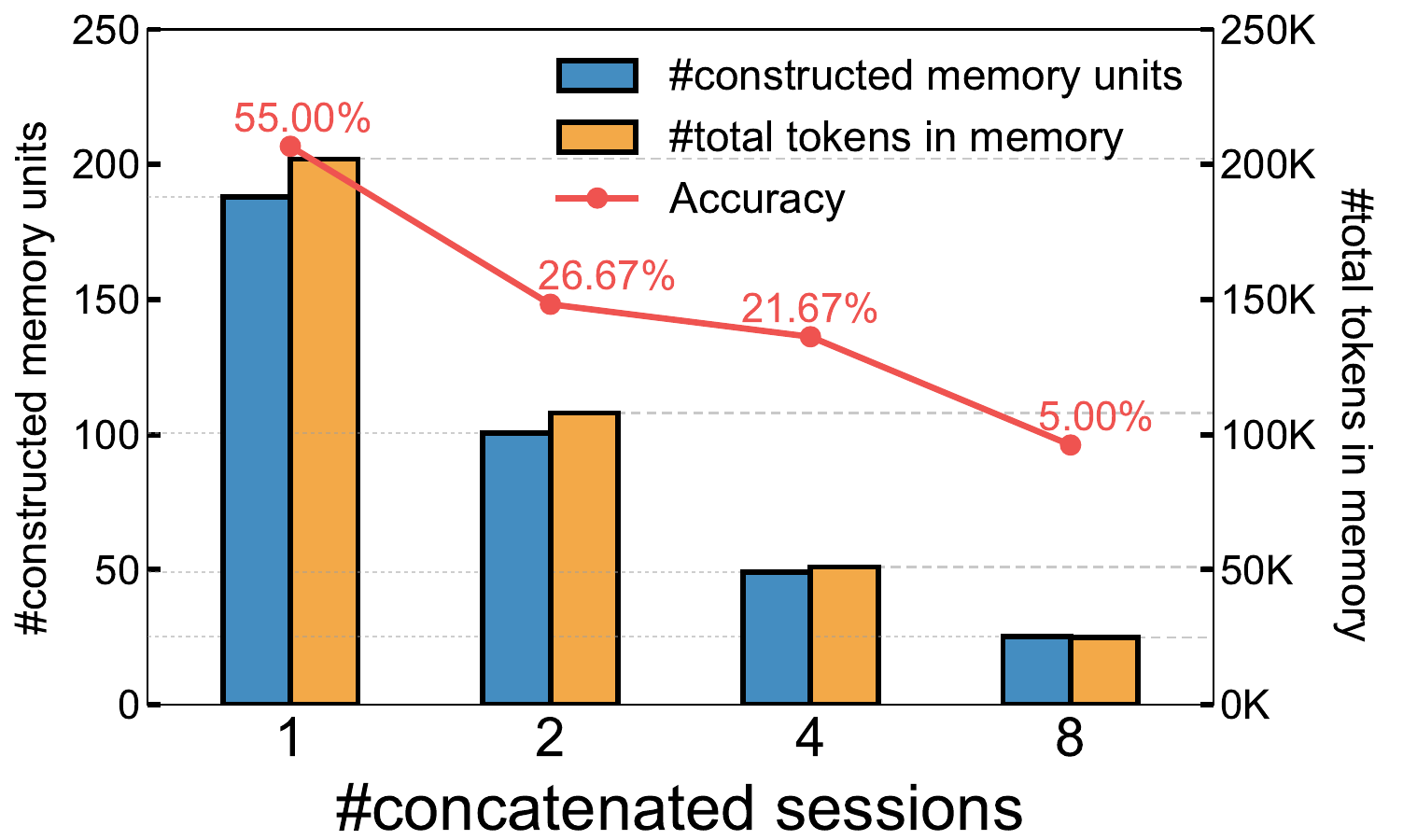}
    \vspace{-1.5em}
    \captionsetup{font=subfigfont,
    margin=0pt,
    justification=centering}
    \caption{Accuracy and memory size of Mem0}
    \label{fig:mem0_memory}
\end{subfigure}\hfill
\begin{subfigure}[t]{0.32\textwidth}
    \centering
    \includegraphics[width=\linewidth]{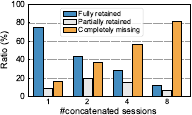}
    \vspace{-1.5em}
    \captionsetup{font=subfigfont,
    margin=0pt,
    justification=centering}
    \caption{Evidence preservation in Mem0}
    \label{fig:mem0_coverage}
\end{subfigure}\hfill
\begin{subfigure}[t]{0.35\textwidth}
    \centering
    \includegraphics[
        width=\linewidth,
    ]{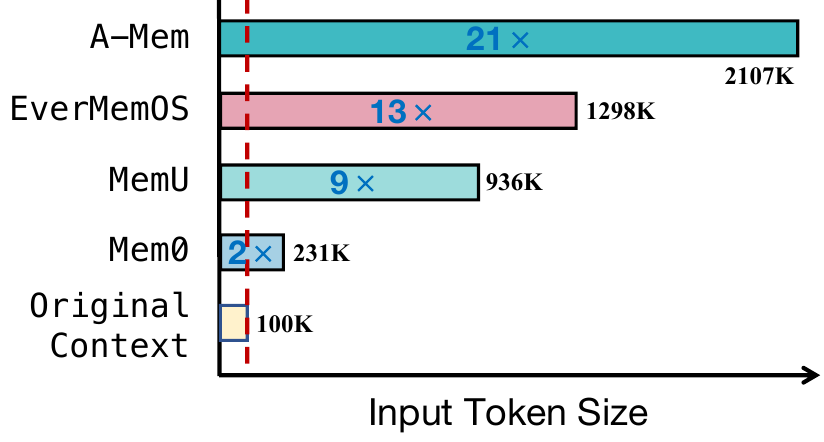}
    \vspace{-1.5em}
    \captionsetup{font=subfigfont,
    margin=0pt,
    justification=centering}
    \caption{Memory construction cost}
    \label{fig:cost}
\end{subfigure}
\vspace{-0.3em}
\caption{Limitations of existing LLM-based memory construction methods. As interaction length and information entanglement increase (simulated via session concatenation with LongMemEval), representative methods exhibit (a) degraded QA accuracy and shrinking memory size, (b) increasing loss of query-relevant evidence, and (c) substantial token amplification during memory construction.}
\label{fig:four_panel}
\vspace{-1.5em}
\end{figure*}

However, we show that this paradigm encounters significant challenges when applied to \textbf{long-horizon, high-entropy} real-world conversational interactions, where dialogues are lengthy, topics are interleaved, and query-relevant evidence may appear sparsely and early in the conversation.
First, \textbf{memory construction becomes lossy and unstable as conversation length and information entropy grow, leading to missing evidence and degraded QA performance.}
While prior methods are typically evaluated on benchmarks such as LongMemEval~\citep{wu2024longmemeval}, these settings significantly underestimate real-world interaction complexity: a single session in LongMemEval contains about 2.1K tokens, whereas real-world LLM usage has already exceeded an average of 5.4K tokens per request~\citep{aubakirova2026state}. 
To bridge this gap, we simulate realistic scenarios by concatenating multiple sessions to construct longer and more entangled interaction histories.
As shown in Figure~\ref{fig:mem0_memory}, our empirical results reveal that as the number of concatenated sessions increases, the representative Mem0~\citep{chhikara2025mem0} method exhibits consistent degradation in downstream QA accuracy. 
This degradation is accompanied by a reduction in both the number of constructed memory units and their information content (quantified by the total number of tokens in their constructed memory units), indicating increasing information loss during memory construction. 
Further analysis in Figure~\ref{fig:mem0_coverage} shows that a substantial portion of query-relevant evidence is either partially retained or completely missing. More details are provided in Appendix~\ref{app:memory analysis}.
Second, \textbf{LLM-based memory construction incurs prohibitive cost}. 
These methods rely heavily on repeated LLM invocation during memory construction, leading to significant token amplification beyond the original input. 
As illustrated in Figure~\ref{fig:cost}, for a 100K-token interaction, the total LLM input tokens during construction can reach 2$\times$, 9$\times$, 13$\times$, and 21$\times$ of the original input for Mem0~\citep{chhikara2025mem0}, MemU\footnote{MemU is an open-source memory system: \url{https://github.com/NevaMind-AI/memU}}, EverMemOS~\citep{hu2026evermemos}, and A-Mem~\citep{xu2025mem}, respectively. 
This substantial overhead not only increases monetary cost but also limits scalability in long-horizon deployments.

These observations suggest that the limitations of existing methods stem from their reliance on \emph{construction-time abstraction}, which is both lossy and expensive under long-horizon, high-entropy interactions. 
\textbf{This motivates a shift in design: instead of transferring interactions into memory during construction, we treat raw interactions as first-class memory with efficient query-time access.} 
This shift avoids construction-time information loss, reduces reliance on repeated LLM invocation, and enables more stable retrieval of query-relevant evidence.
However, this introduces a key challenge: \textbf{how to precisely locate complete query-relevant evidence when it is entangled within long and mixed interaction histories?} In long-horizon, high-entropy settings, information is highly mixed and query-relevant evidence is often mixed with large amounts of irrelevant context, making accurate and complete retrieval difficult. Therefore, this requires organizing interactions in a way that preserves fine-grained evidence while supporting accurate and efficient access at query time.

This paper proposes \texttt{Threader}, a memory system designed for long-horizon, high-entropy conversational agents. \texttt{Threader} eliminates LLM-based abstraction during memory construction. Instead, it organizes interaction logs into \emph{lossless, topic-coherent segments} through a lightweight and incremental segmentation process, preserving fine-grained evidence. It further constructs \emph{multi-view segment representations} that capture semantic, temporal, and structural cues, enabling precise retrieval. At query time, the system performs \emph{multi-signal evidence retrieval} to accurately identify relevant segments by combining segment-level access with localized evidence matching, providing coherent and evidence-complete context for response generation.

We evaluate \texttt{Threader} on multiple long-context conversational benchmarks. Experimental results show that \texttt{Threader} consistently outperforms advanced agent memory systems and agentic RAG methods in answer accuracy and evidence recall. Moreover, it significantly reduces both memory construction overhead and query-time latency, achieving efficient memory access.

Our contributions are summarized as follows: (1) We identify fundamental limitations of LLM-based memory construction in long-horizon, high-entropy conversational settings, supported by empirical analysis on both effectiveness and cost.
(2) We propose \texttt{Threader}, a memory framework that shifts the focus from memory construction to efficient, structure-aware access over raw interactions.
(3) We design a unified system combining incremental segmentation, multi-view representation, and multi-signal retrieval for robust evidence access.
(4) We demonstrate that \texttt{Threader} achieves superior performance while substantially reducing both construction and query-time cost.

\section{Related Work}

\textbf{Agent Memory Paradigms}. Recent taxonomies \citep{hu2025memory} categorize agent memory into three primary architectures: non-parametric (token-level), parametric, and latent memory. Non-parametric memory stores explicit units, such as chunks or summaries, in external datastores. Systems like MemoryBank~\citep{zhong2024memorybank}, MemGPT \citep{packer2023memgpt} and Mem0 \citep{chhikara2025mem0} represent this dominant paradigm, offering interpretability and ease of updates. 
In contrast, parametric memory stores information in model weights, often through editing \citep{wang2021k, meng2022locating, wang2024wise}, whereas latent memory relies on internal representations or recurrent states \citep{wang2024memoryllm, wang2025m+, zhang2025memgen}. However, parametric and latent methods often face challenges in update cost, controllability, and interpretability, making them less suitable for rapidly evolving dialogue memory. Thus, token-level memory remains a natural choice for long-context human--assistant dialogue, where memories must be explicit, updatable, and retrievable.

\textbf{Factual Memory in Dialogue}. Within the non-parametric line, prior work often distinguishes experiential memory, which stores reusable strategies, skills, or trajectories for future task solving \citep{shinn2023reflexion, zhao2024expel, wang2024agent}, from working memory, which maintains transient context for ongoing reasoning within a task \citep{yu2025memagent,wu2025resum,hu2025hiagent}. For sustained human-assistant interaction, however, factual memory—storing user-specific facts and interaction history—remains the central challenge \citep{rasmussen2025zep, kang2025memory,li2025memos, hu2026evermemos}. Most factual-memory systems follow a common pipeline: they rewrite raw dialogue into compact memory units and retrieve them at query time. This reliance on LLM-mediated memory rewriting introduces two practical limitations: memory construction can become increasingly lossy and unstable in realistic long-context dialogue, and the repeated LLM calls required for rewriting impose substantial construction cost.

\textbf{RAG for Memory}. Agent memory differs from classical RAG mainly in persistence: memory stores and reuses information across interactions, rather than only retrieving external knowledge for inference \citep{gao2023retrieval, singh2025agentic}. HippoRAG~\citep{gutierrez2024hipporag} and HippoRAG2~\citep{gutierrez2025rag} connect retrieval with long-term memory, showing that memory can be viewed as structured non-parametric retrieval. Similarly, many factual-memory systems for long-context dialogue store histories in external banks and retrieve relevant units at query time \citep{packer2023memgpt, chhikara2025mem0, rasmussen2025zep, hu2026evermemos}. Unlike conventional RAG, conversational memory must first organize evolving interactions into retrievable units. We therefore treat memory formation not as semantic rewriting, but as evidence-preserving structural organization for efficient access.

\section{Methodology}
\label{method}
\begin{figure*}[!tb]
\centering
\includegraphics[width=1\linewidth]{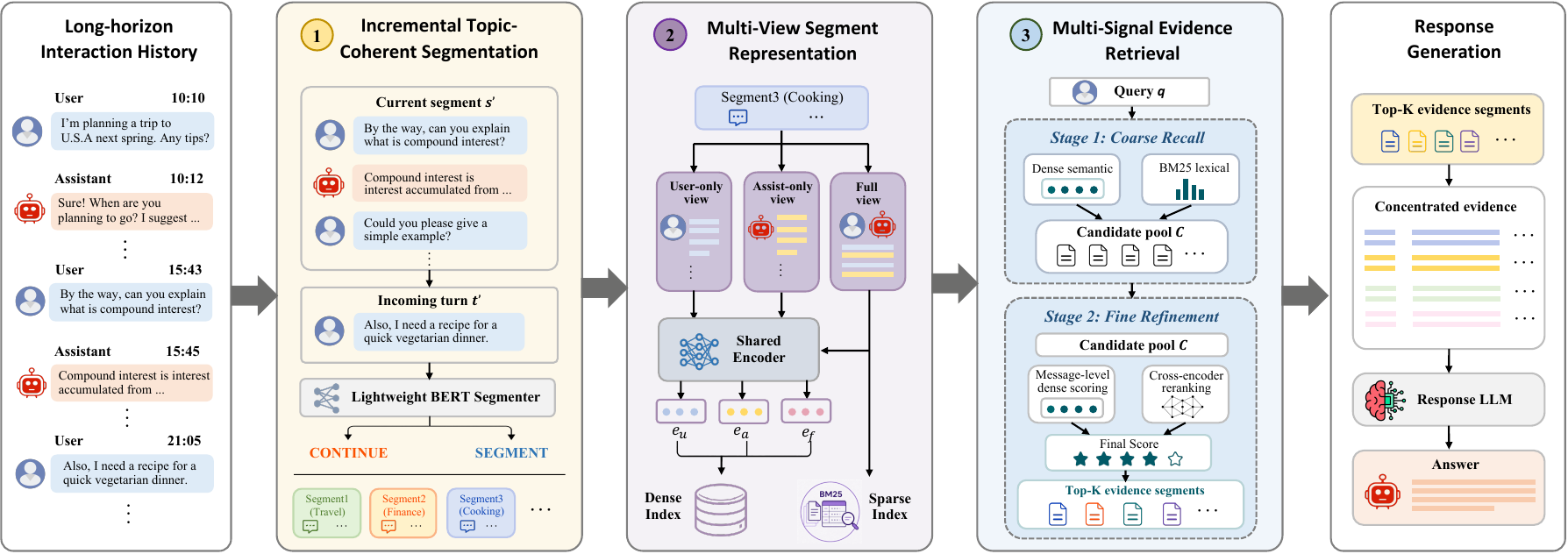}
\vspace{-0.8em}
\caption{\textbf{Overview of \texttt{Threader}.} It incrementally segments interaction streams into topic-coherent units, encodes them with multi-view representations, and retrieves related evidence with multi-signals.}
\label{fig:method_overview}
\vspace{-1.2em}
\end{figure*}

As illustrated in Fig.~\ref{fig:method_overview}, \texttt{Threader} makes raw interaction histories usable as long-term memory through structure-aware access rather than LLM-based rewriting. 
This requires solving three challenges: identifying coherent evidence boundaries, representing heterogeneous evidence, and retrieving complete evidence from long and entangled histories. 
Accordingly, \texttt{Threader} consists of three components. 
\textbf{(1) Incremental Topic-Coherent Segmentation} partitions interaction histories into semantically consistent units while preserving the original fine-grained evidence. 
\textbf{(2) Multi-View Segment Representation} encodes each segment from complementary perspectives to avoid collapsing user-side, assistant-side, and full-context evidence into a single representation. 
\textbf{(3) Multi-Signal Evidence Retrieval} combines dense, lexical, and reranking signals to locate query-relevant evidence and reconstructs the retrieved segments into temporally consistent context for response generation.

\subsection{Incremental Topic-Coherent Segmentation}
\label{sec:segmenter}
In long-horizon conversational settings, interaction histories are often \emph{high-entropy}, where multiple topics are interleaved and query-relevant evidence is sparsely distributed. 
A common practice in RAG
systems is to partition text into fixed-size chunks. 
However, such chunking is agnostic to semantic boundaries, often fragmenting coherent evidence or mixing unrelated content, which hinders precise retrieval. 
Instead, we aim to segment interactions into \emph{topic-coherent evidence segments}, such that each segment forms a semantically consistent unit that can be retrieved as a whole.

Let a conversation be denoted as $C=\{S_1,\dots,S_n\}$, where each session $S_i$ is an ordered sequence of turns:
$S_i = (t_1, t_2, \dots, t_m)$. Our goal is to partition each session into a sequence of contiguous, topic-coherent segments: $\mathcal{S}_i = (s_1, s_2, \dots, s_K),
s_k = (t_{k,1}, t_{k,2}, \dots, t_{k,n_k}),$ where each segment $s_k$ is a contiguous subsequence of turns sharing a consistent semantic topic.

An effective segmenter for conversational memory should satisfy two key properties:
1) \textbf{Lightweight}, to avoid the high cost of LLM-based preprocessing; 
2) \textbf{Incremental}, to support streaming interactions without requiring full-context recomputation.
Prior work has explored topic-aware segmentation using LLMs (e.g., SECOM \citep{pan2025memory}, EverMemOS \citep{hu2026evermemos}), but such approaches incur substantial computational overhead. 
More lightweight methods (e.g., LightMem \citep{fang2025lightmem}) rely on embedding or attention-based heuristics, but are typically not fully incremental, as they require buffering and batch processing before determining segment boundaries.

To achieve both efficiency and incrementality, we formulate segmentation as a \emph{binary decision problem} over streaming turns. 
The system maintains a current segment $s'$ (temporal history), and processes incoming turns sequentially. 
Given a new turn $t'$, the model predicts whether $t'$ should be appended to $s'$ or start a new segment: \textsc{CONTINUE}: append $t'$ to the current segment;
\textsc{SEGMENT}: finalize the current segment and start a new one with $t'$.
Formally, let $s'=(t'_1,\dots,t'_j)$ denote the current segment. 
For an incoming turn $t'_{j+1}$, we construct an input: $x = \mathrm{Concat}(s', [\mathrm{SEP}], t'_{j+1})$, and predict a binary label $y \in \{0,1\}$ indicating \textsc{CONTINUE} or \textsc{SEGMENT}. 
This formulation enables fully online segmentation without revisiting past context.

We instantiate the segmenter using a lightweight BERT encoder~\citep{devlin2019bert}, which provides strong bidirectional semantic modeling while maintaining low computational overhead. We construct training labels using both coarse-grained and fine-grained signals. 
Session boundaries from LongMemEval~\citep{wu2024longmemeval} are used as \emph{hard labels}, capturing clear conversational shifts. 
To further refine intra-session topic boundaries, we follow prior work and leverage LLM-based annotation to obtain \emph{soft labels}~\citep{pan2025memory}. 
This combination allows the model to capture both session-level transitions and finer-grained topic changes.

Given segmented conversations, we construct binary classification examples for incremental decision making. 
For turns within the same segment, we create \textsc{CONTINUE} examples:
\begin{equation}
\left(
x^{\mathrm{cont}}_{k,j},\, y^{\mathrm{cont}}_{k,j}
\right)
=
\left(
\mathrm{Concat}((t_{k,1},\dots,t_{k,j}),[\mathrm{SEP}],t_{k,j+1}),\,0
\right), \quad j=1,\dots,n_{k-1}
\label{eq:continue}
\end{equation}
For segment boundaries, we create \textsc{SEGMENT} examples:
\begin{equation}
\left(
x^{\mathrm{seg}}_{k},\, y^{\mathrm{seg}}_{k}
\right)
=
\left(
\mathrm{Concat}((t_{k,1},\dots,t_{k,n_k}),[\mathrm{SEP}],t_{k+1,1}),\,1
\right), \quad k=1,\dots,K-1
\label{eq:segment}
\end{equation}
The model is trained using the standard cross-entropy loss. Full details are provided in Appendix~\ref{app:segmenter_training}.

\subsection{Multi-View Segment Representation}
A fundamental challenge in long-horizon conversational retrieval is that evidence is \emph{heterogeneous}: 
user utterances often encode persistent and query-relevant facts (e.g., preferences, experiences), 
while assistant responses are typically longer, more verbose, and contain substantial generative content. 
A naive approach is to encode each segment into a single embedding. 
However, this inevitably collapses heterogeneous evidence into a unified representation, where dominant assistant-side content can overshadow weak but critical user-side signals, degrading retrieval of query-relevant evidence.

To address this, we adopt a \emph{multi-view representation} that explicitly separates different evidence sources. 
Given a segment $s_i=(t_{i,1},\dots,t_{i,m})$ with turns $t_{i,j}=(u_{i,j},a_{i,j})$, 
we construct three complementary views:
$
x_i^{(u)} = \mathrm{Concat}(u_{i,1},\dots,u_{i,m}), 
x_i^{(a)} = \mathrm{Concat}(a_{i,1},\dots,a_{i,m}), 
x_i^{(f)} = \mathrm{Concat}(u_{i,1},a_{i,1},\dots,u_{i,m},a_{i,m}),
$
corresponding to user-only, assistant-only, and full interaction representations. 
Each view is encoded using a shared encoder $f_{\mathrm{enc}}$: $e_i^{(v)} = f_{\mathrm{enc}}(x_i^{(v)}),  v \in \{u,a,f\}$.
This design decouples evidence into complementary representations, 
allowing retrieval to selectively attend to user-centric signals while preserving contextual completeness.

\subsection{Multi-Signal Evidence Retrieval}
\label{method:retrieve}

Given a query $q$ and segments $\{s_i\}$, we retrieve segments that jointly maximize evidence relevance and completeness. 
We formulate this as a \emph{multi-signal relevance estimation problem}, where different signals capture complementary aspects of evidence. Specifically, we consider three retrieval signals.

Given the query embedding $z(q)=f_{\mathrm{enc}}(q)$, the
\textbf{dense semantic signal} aggregates similarities across
the three segment views:
\begin{equation}
\mathrm{score}^{(e)}_i(q)
=
\sum_{v \in \{u,a,f\}} w_v
\cos\!\big(z(q), e_i^{(v)}\big),
\quad \sum_v w_v = 1.
\label{eq:dense}
\end{equation}
We complement this with a \textbf{lexical matching signal},
$\mathrm{score}^{(b)}_i(q)=\mathrm{BM25}(q,s_i)$,
and a \textbf{cross-encoder signal},
$\mathrm{score}^{(r)}_i(q)=\mathrm{Rerank}(q,s_i)$.
Dense and lexical scores support efficient candidate retrieval,
while the cross-encoder captures fine-grained query--segment
interactions at higher computational cost, motivating a
\emph{coarse-to-fine two-stage retrieval strategy}.

\textbf{Stage I: High-recall segment retrieval.}
We first construct a high-recall candidate pool $\mathcal{C}$,
which provides a shortlist of segments for fine-grained relevance
estimation in Stage~II. Specifically, we take the union of the
top-$K_1$ segments retrieved by dense retrieval and BM25:
\begin{equation}
\mathcal{C}
=
\mathrm{Top}\text{-}K_1(\mathrm{score}^{(e)}(q))
\;\cup\;
\mathrm{Top}\text{-}K_1(\mathrm{score}^{(b)}(q)).
\label{eq:pool}
\end{equation}
This combines semantic coverage with lexical matching to improve recall.

\textbf{Stage II: Evidence-level refinement.}
A key limitation of segment-level retrieval is that query-relevant evidence is often \emph{locally concentrated within segments}, rather than uniformly distributed across all turns. 
To capture such fine-grained signals, we refine dense matching from segment-level to message-level scoring.

Concretely, let candidate segment $s_i$ contain $m_i$ turns, denoted as 
$\{t_{i,j}=(u_{i,j}, a_{i,j})\}_{j=1}^{m_i}$.
For each turn, we compute the similarity between the query and each user/assistant message using the encoder $f_{\mathrm{enc}}$; message embeddings are pre-computed during indexing.
This produces a set of message-level relevance scores within the segment. 
We then select the Top-$K_2$ highest scores and average them to obtain a refined dense score:
\begin{equation}
\widetilde{\mathrm{score}}^{(e)}_i(q)
=
\frac{1}{K_2}
\sum_{k=1}^{K_2}
\mathrm{TopK}_k\big(
\{\cos(z(q), f_{\mathrm{enc}}(u_{i,j})), 
\cos(z(q), f_{\mathrm{enc}}(a_{i,j}))\}_{j=1}^{m_i}
\big).
\label{eq:msg}
\end{equation}
Before each fusion step, scores are min-max normalized over candidate segments for comparable scales.
We then combine the normalized dense signal with lexical matching:
\begin{equation}
\mathrm{score}^{(h)}_i
=
\widetilde{\mathrm{score}}^{(e)}_i(q)
+
\beta\,\mathrm{score}^{(b)}_i(q),
\label{eq:hybrid}
\end{equation}
and further apply cross-encoder reranking to obtain the final score:
\begin{equation}
\mathrm{score}_i
=
\alpha\,\mathrm{score}^{(h)}_i
+
(1-\alpha)\,\mathrm{score}^{(r)}_i(q).
\label{eq:final}
\end{equation}

From the final score $\mathrm{score}_i$, we select the top-$K$ segments for downstream response generation. In particular, although Stage II operates at message-level granularity, we ultimately perform retrieval at the segment level. 
This design preserves \emph{discourse coherence} in downstream reasoning, as query-relevant evidence in conversational agents is often meaningful only within its surrounding context. 
In this way, fine-grained scoring is used to identify evidence within segments, while segment-level retrieval ensures that retrieved content remains structurally and temporally consistent for generation.

\section{Experiments}
\label{sec:experiments}
\subsection{Experimental Setup}
\label{sec:setup}
\textbf{Datasets and Evaluation.}
We evaluate on \textbf{LongMemEval}~\citep{wu2024longmemeval} and \textbf{PersonaMem}~\citep{jiang2025know}, two benchmarks that capture complementary long-term memory settings. 
LongMemEval contains 500 QA instances with an average session length of 2K tokens, covering information extraction, multi-session reasoning, temporal reasoning, and knowledge updates. 
PersonaMem provides a more challenging personalization setting with substantially longer sessions (589 QA instances, 26K tokens on average), emphasizing long-horizon user-specific memory.
Dataset details are provided in Appendix~\ref{app:datasets}. We evaluate LongMemEval using \textbf{LLM-as-a-Judge} and PersonaMem using exact match. Detailed judge protocols are provided in Appendix~\ref{app:judge}.

\textbf{Baselines.}
We compare \texttt{Threader} with seven representative baselines, including five \emph{agent memory} methods and two \emph{agentic RAG} approaches.
The memory baselines include \textbf{Mem0}~\citep{chhikara2025mem0}, \textbf{MemU}, \textbf{A-Mem}~\citep{xu2025mem}, \textbf{MemOS}~\citep{li2025memos}, and \textbf{EverMemOS}~\citep{hu2026evermemos}, covering recent advances in memory construction, organization, updating, and retrieval.
The agentic RAG baselines include \textbf{Self-RAG}~\citep{asai2024selfrag} and \textbf{HippoRAG2}~\citep{gutierrez2025rag}, which represent strong retrieval-based agents without explicit memory construction. 
Details of the baselines are available in Appendix~\ref{app:implement}.

\textbf{Implementation Details.}
Following our lightweight design principle, \texttt{Threader} uses paraphrase-multilingual-MiniLM-L12-v2~\citep{reimers-2019-sentence-bert} for embedding and bge-reranker-base~\citep{bge_embedding} for reranking.
Our main setting uses GPT-4.1-mini for memory construction and GPT-5-mini for response generation, reflecting practical deployments that prioritize cost-efficient construction and high-quality reasoning.
We also report a low-latency setting with GPT-4.1-mini and GPT-5-nano.
Results with Qwen models are provided in Appendix~\ref{app:qwen}.
For fair comparison, all LLM-based construction methods share the same construction model, and all methods share the same response model while retaining method-specific components.

\subsection{Main Results}

Table~\ref{tab:longmemeval_main} and Table~\ref{tab:personamem_main} report the main results. 
Overall, \texttt{Threader} consistently outperforms all baselines across both datasets and response backbones, demonstrating the effectiveness of structure-aware access over LLM-based memory construction or retrieval in long-horizon conversational settings.
On LongMemEval, \texttt{Threader} improves over the strongest baseline by 3.33\% with GPT-5-nano and 6.84\% with GPT-5-mini. On PersonaMem, \texttt{Threader} also obtains the best overall score, with relative gains of 4.50\% and 7.29\%, respectively.

On LongMemEval, the performance gains are especially pronounced on evidence-intensive question types. \texttt{Threader} improves \textit{Temporal-Reasoning} by 6.94\% with GPT-5-nano and 11.02\% with GPT-5-mini, and improves \textit{Knowledge-Update} by 5.80\% with GPT-5-mini. It also achieves competitive results on \textit{Multi-Session}, \textit{Single-Session-User}, and \textit{Single-Session-Assistant} tasks. These tasks require locating sparse and temporally distant evidence, where lossy memory construction often fails. 
In contrast, \texttt{Threader} preserves fine-grained interaction structure and retrieves topic-coherent segments, leading to more complete evidence coverage. We observe a similar trend on PersonaMem, where \texttt{Threader} achieves the largest improvements on \textit{generalize-to-new-scenarios} and \textit{recalling-user-shared-facts}, further validating its advantage in long and entangled conversational histories.

\begin{table*}[t]
\caption{\textbf{Main results on LongMemEval.} \texttt{Threader} retrieves the top-12 segments for LongMemEval. SS-User, SS-Assi., SS-Pref., Multi-S., Know.-U., and Temp.-R. denote Single-Session-User, Single-Session-Assistant, Single-Session-Preference, Multi-Session, Knowledge-Update, and Temporal-Reasoning, respectively. Results are grouped by backbone. Within each backbone block, the best result in each column is shown in \textbf{bold}, the second-best distinct result is \underline{underlined}.}
\label{tab:longmemeval_main}
\centering
\footnotesize
\renewcommand{\arraystretch}{1.0}
\setlength{\tabcolsep}{1.8pt}{
\begin{tabular}{@{}lccccccc@{}}
\toprule
\textbf{Method} & \textbf{SS-User} & \textbf{SS-Assi.} & \textbf{SS-Pref.} & \textbf{Multi-S.} & \textbf{Know.-U.} & \textbf{Temp-R.} & \textbf{Overall} \\
\midrule

\multicolumn{8}{c}{\textit{GPT-5-nano}} \\
Self-RAG     & 87.14 & \underline{94.64} & 43.33 & 56.39 & 73.08 & 73.68 & 71.40 \\
HippoRAG2    & 71.43 & \underline{94.64} & 53.33 & 60.15 & 76.92 & 69.92 & 70.40 \\
Mem0         & 80.00 & 46.43 & 26.67 & 39.85 & 69.23 & 58.65 & 55.00 \\
MemU         & 81.43 & 78.57 & 43.33 & 55.64 & 53.85 & 68.42 & 64.20 \\
A-Mem        & 84.29 & 82.14 & 20.00 & 30.83 & 56.41 & 27.82 & 46.60 \\
MemOS        & 81.43 & 87.50 & \textbf{100.00} & 57.89 & 71.79 & 45.11 & 65.80 \\
EverMemOS    & \underline{92.85} & 78.57 & \underline{90.00} & \underline{66.92} & \underline{82.05} & \underline{75.93} & \underline{78.00} \\
\rowcolor{gray!25}
\textbf{\texttt{Threader}} & \textbf{94.29}\,\mbox{\tiny($\uparrow$1.55\%)} & \textbf{98.21}\,\mbox{\tiny($\uparrow$3.77\%)} & 63.33\,\mbox{\tiny($\downarrow$36.67\%)} & \textbf{67.67}\,\mbox{\tiny($\uparrow$1.12\%)} & \textbf{83.33}\,\mbox{\tiny($\uparrow$1.56\%)} & \textbf{81.20}\,\mbox{\tiny($\uparrow$6.94\%)} & \textbf{80.60}\,\mbox{\tiny($\uparrow$3.33\%)} \\
\midrule

\multicolumn{8}{c}{\textit{GPT-5-mini}} \\
Self-RAG     & 92.86 & \underline{96.43} & 60.00 & 61.65 & 78.21 & 77.44 & 76.60 \\
HippoRAG2    & 72.86 & 92.86 & 70.00 & 67.67 & 87.18 & 69.92 & 75.00 \\
Mem0         & 81.43 & 42.86 & 26.67 & 44.36 & 71.79 & 61.65 & 57.20 \\
MemU         & 87.14 & 82.14 & 40.00 & 56.39 & 79.49 & 76.69 & 71.60 \\
A-Mem        & 88.57 & \textbf{98.21} & 43.33 & 51.13 & 76.92 & 58.65 & 67.20 \\
MemOS        & 82.86 & 89.29 & \textbf{100.00} & 60.90 & 79.49 & 45.86 & 68.40 \\
EverMemOS    & \underline{95.71} & 80.35 & \underline{96.67} & \underline{78.95} & \underline{88.46} & \underline{81.95} & \underline{84.80} \\
\rowcolor{gray!25}
\textbf{\texttt{Threader}} & \textbf{98.57}\,\mbox{\tiny($\uparrow$2.99\%)} & \textbf{98.21}\,\mbox{\tiny($\rightarrow$0.00\%)} & \underline{96.67}\,\mbox{\tiny($\downarrow$3.33\%)} & \textbf{79.70}\,\mbox{\tiny($\uparrow$0.95\%)} & \textbf{93.59}\,\mbox{\tiny($\uparrow$5.80\%)} & \textbf{90.98}\,\mbox{\tiny($\uparrow$11.02\%)} & \textbf{90.60}\,\mbox{\tiny($\uparrow$6.84\%)} \\
\bottomrule
\end{tabular}
}
\vspace{-1.0em}
\end{table*}

\newcommand{\vtiny}{\fontsize{4.7pt}{5pt}\selectfont}

\begin{table*}[tb!]
\caption{\textbf{Main results on PersonaMem.} \texttt{Threader} retrieves the top-8 segments for PersonaMem. Generalize-New-Scn., Preference-Align, User-Shared, User-Mentioned, Update-Reason, New-Ideas, and Preference-Evolution denote generalizing-to-new-scenarios, provide-preference-aligned-recommendations, recalling-user-shared-facts, recalling-facts-mentioned-by-the-user, recalling-reasons-behind-previous-updates, suggesting-new-ideas, and tracking-full-preference-evolution, respectively. Results are grouped by backbone. Within each backbone block, the best result in each column is shown in \textbf{bold}, the second-best distinct result is \underline{underlined}.}
\label{tab:personamem_main}
\centering
\footnotesize
\renewcommand{\arraystretch}{1.0}

\setlength{\tabcolsep}{1pt}{

\begin{tabular}{@{}lcccccccc@{}}
\toprule
\multirow{2}{*}{\textbf{Method}} 
& \textbf{Generalize} 
& \textbf{Preference} 
& \textbf{User-} 
& \textbf{User-} 
& \textbf{Update-} 
& \textbf{New-} 
& \textbf{Preference} 
& \multirow{2}{*}{\textbf{Overall}} \\
& \textbf{-New-Scn.} 
& \textbf{-Align} 
& \textbf{Shared} 
& \textbf{Mentioned} 
& \textbf{Reason} 
& \textbf{Ideas} 
& \textbf{Evolution} 
& \\
\midrule

\multicolumn{9}{c}{\textit{GPT-5-nano}} \\
Self-RAG     & 29.82 & 52.73 & 16.28 & 47.06 & 67.68 & \underline{38.71} & 48.20 & 41.60 \\
HippoRAG2    & 56.14 & 65.45 & 53.49 & \textbf{70.59} & 69.70 & \textbf{39.78} & 51.08 & 55.35 \\
Mem0         & 45.61 & 43.64 & 31.01 & 47.06 & 70.71 & \textbf{39.78} & \underline{55.40} & 47.88 \\
MemU         & 35.09 & 41.82 & 47.29 & 41.18 & 58.59 & 27.96 & 51.80 & 45.33 \\
A-Mem        & 49.12 & 67.27 & 60.47 & 52.94 & 63.64 & 29.03 & 30.94 & 48.39 \\
MemOS        & \underline{63.16} & \underline{69.09} & \underline{68.22} & 58.82 & \underline{76.77} & 34.41 & 54.68 & \underline{60.44} \\
EverMemOS    & 50.88 & 61.82 & 58.91 & \underline{64.71} & \textbf{77.78} & \underline{38.71} & 51.80 & 56.88 \\
\rowcolor{gray!25}
\textbf{\texttt{Threader}} & \textbf{78.95}\,\mbox{\vtiny($\uparrow$25.00\%)} & \textbf{72.73}\,\mbox{\vtiny($\uparrow$5.27\%)} & \textbf{69.77}\,\mbox{\vtiny($\uparrow$2.27\%)} & 52.94\,\mbox{\vtiny($\downarrow$25.00\%)} & 72.73\,\mbox{\vtiny($\downarrow$6.49\%)} & 37.63\,\mbox{\vtiny($\downarrow$5.40\%)} & \textbf{58.27}\,\mbox{\vtiny($\uparrow$5.18\%)} & \textbf{63.16}\,\mbox{\vtiny($\uparrow$4.50\%)} \\
\midrule

\multicolumn{9}{c}{\textit{GPT-5-mini}} \\
Self-RAG     & 29.82 & 52.73 & 4.65 & 58.82 & 55.56 & \textbf{48.39} & 54.68 & 40.41 \\
HippoRAG2    & 61.40 & 63.64 & 68.22 & \textbf{70.59} & 67.68 & 37.63 & 56.12 & 59.42 \\
Mem0         & 57.89 & 41.82 & 29.46 & 35.29 & 68.69 & 27.96 & \textbf{66.91} & 48.73 \\
MemU         & 54.39 & 41.82 & 60.47 & 52.94 & 67.68 & 27.96 & 61.87 & 54.33 \\
A-Mem        & 70.18 & 65.45 & 70.54 & \underline{64.71} & 72.73 & 29.03 & 58.27 & 60.78 \\
MemOS        & \underline{71.93} & \textbf{72.73} & 72.87 & 58.82 & \underline{82.83} & 30.11 & 61.87 & \underline{64.69} \\
EverMemOS    & 68.42 & 67.27 & \underline{74.42} & \textbf{70.59} & \textbf{86.87} & 26.88 & 64.03 & \underline{65.20} \\
\rowcolor{gray!25}
\textbf{\texttt{Threader}} & \textbf{84.21}\,\mbox{\vtiny($\uparrow$17.07\%)} & \underline{70.91}\,\mbox{\vtiny($\downarrow$2.50\%)} & \textbf{80.62}\,\mbox{\vtiny($\uparrow$8.33\%)} & \textbf{70.59}\,\mbox{\vtiny($\rightarrow$0.00\%)} & 79.80\,\mbox{\vtiny($\downarrow$8.14\%)} & \underline{45.16}\,\mbox{\vtiny($\downarrow$6.67\%)} & 63.31\,\mbox{\vtiny($\downarrow$5.38\%)} & \textbf{69.95}\,\mbox{\vtiny($\uparrow$7.29\%)} \\
\bottomrule
\end{tabular}
}
\vspace{-1.5em}
\end{table*}

\texttt{Threader} is less advantageous on tasks such as Single-Session-Preference (LongMemEval) and \textit{tracking-full-preference-evolution} (PersonaMem), which require aggregating and abstracting user preferences across interactions. 
Unlike memory-construction approaches that explicitly summarize preferences, \texttt{Threader} operates on raw evidence and relies on the response model for higher-level synthesis. This suggests a complementary trade-off between evidence preservation and abstraction.

Overall, the results indicate that \texttt{Threader} is particularly effective when accurate memory access depends on preserving and retrieving distributed evidence, which is the dominant challenge in long-horizon, high-entropy conversational settings.

\subsection{Ablation Studies}

\newcolumntype{M}{>{\centering\arraybackslash}p{0.70cm}}

\begin{table*}[t]
\centering
\scriptsize
\setlength{\tabcolsep}{2.6pt}

\caption{\textbf{Ablation study on the key design axes of \texttt{Threader}.}
We evaluate the impact of segmentation, multi-view representation, multi-signal retrieval, and retrieval granularity on answer accuracy and evidence recall under Top-5 and Top-12 settings. 
Full-R denotes the fraction of queries with complete evidence coverage, and Mean-R denotes average evidence recall. 
Emb.-User, Emb.-Asst., and Emb.-Full correspond to using only user, assistant, or full interaction views, respectively.}
\label{tab:ablation_main}

\resizebox{\textwidth}{!}{
\begin{tabular}{@{}c@{\hspace{1.5mm}}|@{\hspace{1.5mm}}c@{}}

\begingroup
\renewcommand{\arraystretch}{1.077}
\begin{tabular}{lMMMMMM}
\toprule
\multirow{2}{*}{\textbf{Variant}} 
& \multicolumn{3}{c}{\textbf{Top-5}} 
& \multicolumn{3}{c}{\textbf{Top-12}} \\
\cmidrule(lr){2-4} \cmidrule(lr){5-7}
& {\tiny \mbox{ACC}}
& {\tiny \mbox{Full-R}}
& {\tiny \mbox{Mean-R}}
& {\tiny \mbox{ACC}}
& {\tiny \mbox{Full-R}}
& {\tiny \mbox{Mean-R}} \\
\midrule

\multicolumn{7}{l}{\textcolor{teal}{\textit{\textbf{Segmentation ablation}}}} \\

\cellcolor{gray!25}\texttt{Threader}
& \cellcolor{gray!25}82.60
& \cellcolor{gray!25}82.55
& \cellcolor{gray!25}90.26
& \cellcolor{gray!25}90.60
& \cellcolor{gray!25}93.40
& \cellcolor{gray!25}96.50 \\

w/o Seg.
& 82.00 & -- & -- & 87.00 & -- & -- \\

Chunk-500
& 82.40 & -- & -- & 85.60 & -- & -- \\

Chunk-1K
& 74.20 & -- & -- & 86.40 & -- & -- \\

Chunk-2K
& 63.60 & -- & -- & 79.80 & -- & -- \\

\cmidrule(lr){1-7}

\multicolumn{7}{l}{\textcolor{teal}{\textit{\textbf{Multi-view representation ablation}}}} \\

\cellcolor{gray!25}Emb. w/o Msg.
& \cellcolor{gray!25}71.20
& \cellcolor{gray!25}67.45
& \cellcolor{gray!25}77.34
& \cellcolor{gray!25}85.40
& \cellcolor{gray!25}86.81
& \cellcolor{gray!25}91.65 \\

Emb. - Full
& 70.20 & 64.68 & 75.52 & 84.00 & 82.55 & 88.72 \\

Emb. - User
& 68.20 & 61.91 & 73.45 & 83.00 & 82.76 & 89.07 \\

Emb. - Asst.
& 62.40 & 55.11 & 68.59 & 75.40 & 73.40 & 82.88 \\

\bottomrule
\end{tabular}
\endgroup

&

\begingroup
\renewcommand{\arraystretch}{1.000}
\begin{tabular}{lMMMMMM}
\toprule
\multirow{2}{*}{\textbf{Variant}} 
& \multicolumn{3}{c}{\textbf{Top-5}} 
& \multicolumn{3}{c}{\textbf{Top-12}} \\
\cmidrule(lr){2-4} \cmidrule(lr){5-7}
& {\tiny \mbox{ACC}}
& {\tiny \mbox{Full-R}}
& {\tiny \mbox{Mean-R}}
& {\tiny \mbox{ACC}}
& {\tiny \mbox{Full-R}}
& {\tiny \mbox{Mean-R}} \\
\midrule

\multicolumn{7}{l}{\textcolor{teal}{\textit{\textbf{Retrieval signal ablation}}}} \\

\cellcolor{gray!25}\texttt{Threader}
& \cellcolor{gray!25}82.60
& \cellcolor{gray!25}82.55
& \cellcolor{gray!25}90.26
& \cellcolor{gray!25}90.60
& \cellcolor{gray!25}93.40
& \cellcolor{gray!25}96.50 \\

w/o Emb.
& 79.00 & 79.15 & 87.40 & 84.60 & 86.17 & 92.19 \\

w/o BM25
& 81.60 & 80.43 & 88.21 & 88.20 & 91.49 & 94.81 \\

w/o Rerank.
& 78.40 & 77.23 & 86.43 & 88.20 & 91.49 & 95.08 \\

w/o Emb.\&BM25
& 80.40 & 76.17 & 85.31 & 85.60 & 88.51 & 93.54 \\

w/o Emb.\&Rerank.
& 74.80 & 73.19 & 81.83 & 83.20 & 84.68 & 91.06 \\

w/o BM25\&Rerank.
& 72.00 & 69.57 & 80.63 & 87.00 & 89.15 & 93.76 \\

w/o Msg.-Emb.
& 81.80 & 81.28 & 88.72 & 89.20 & 93.19 & 96.29 \\

\cmidrule(lr){1-7}

\multicolumn{7}{l}{\textcolor{teal}{\textit{\textbf{Retrieval granularity ablation}}}} \\

\cellcolor{gray!25}Segment Unit
& \cellcolor{gray!25}82.60
& \cellcolor{gray!25}82.55
& \cellcolor{gray!25}90.26
& \cellcolor{gray!25}90.60
& \cellcolor{gray!25}93.40
& \cellcolor{gray!25}96.50 \\

Message Unit
& 82.00 & -- & -- & 85.80 & -- & -- \\

\bottomrule
\end{tabular}
\endgroup

\end{tabular}
}

\vspace{-1.5em}
\end{table*}

As shown in Table~\ref{tab:ablation_main}, we conduct a comprehensive ablation study to evaluate the key design axes of \texttt{Threader}, including 
(i) topic-coherent segmentation, 
(ii) multi-view segment representation, 
(iii) multi-signal retrieval, and 
(iv) retrieval granularity (segment-level in \texttt{Threader} vs. message-level). 
These ablations are designed to test how different components contribute to robust information access under entangled conversational settings.

\textbf{Segmentation ablation.}
Disabling segmentation or replacing topic-coherent segmentation with fixed-size chunking consistently degrades performance, suggesting that retrieval quality depends on semantically aligned evidence units rather than coverage alone.
Coherent segments provide more faithful boundaries for evidence, reducing fragmentation and improving downstream matching.
\textbf{Multi-view representation ablation.}
We isolate multi-view representation under dense-only retrieval by disabling message-level reranking.
Removing views from the multi-view encoding leads to noticeable performance drops. 
This indicates that collapsing heterogeneous interaction signals into a single representation harms retrieval, as weaker but important user-side evidence can be overshadowed. 
Multi-view representations mitigate this effect by separating complementary information sources.
\textbf{Multi-signal retrieval ablation.}
We observe that eliminating individual retrieval signals
degrades performance, while removing their combination leads to larger drops. 
This confirms that different signals capture complementary aspects of relevance—semantic similarity, lexical matching, and fine-grained interaction—and their integration is critical for robust retrieval.
\textbf{Retrieval granularity ablation.}
To evaluate the effectiveness of the segment-level retrieval strategy adopted in \texttt{Threader}, we compare with message-level retrieval (i.e., treating each user or assistant message as an independent retrieval unit instead of full segments).
The results show that, while fine-grained message-level matching improves the ability to locate relevant signals, retrieving individual messages alone degrades performance. 
This indicates that although relevant information is often localized, it typically relies on surrounding context for correct interpretation. 
Segment-level retrieval provides a better trade-off by preserving contextual coherence while enabling fine-grained evidence localization through message-level scoring.
\textbf{Overall}, these results demonstrate that the effectiveness of \texttt{Threader} arises from the joint design of structure-aware segmentation, multi-view representation, and multi-signal retrieval, which together enable accurate and stable access to relevant information.

\subsection{Hyperparameter Analysis}
\label{sec:hyper}

\begin{wrapfigure}{r}{0.5\linewidth}
    \vspace{-2em}
    \centering
    \begin{subfigure}[t]{0.48\linewidth}
        \centering
        \includegraphics[width=\linewidth]{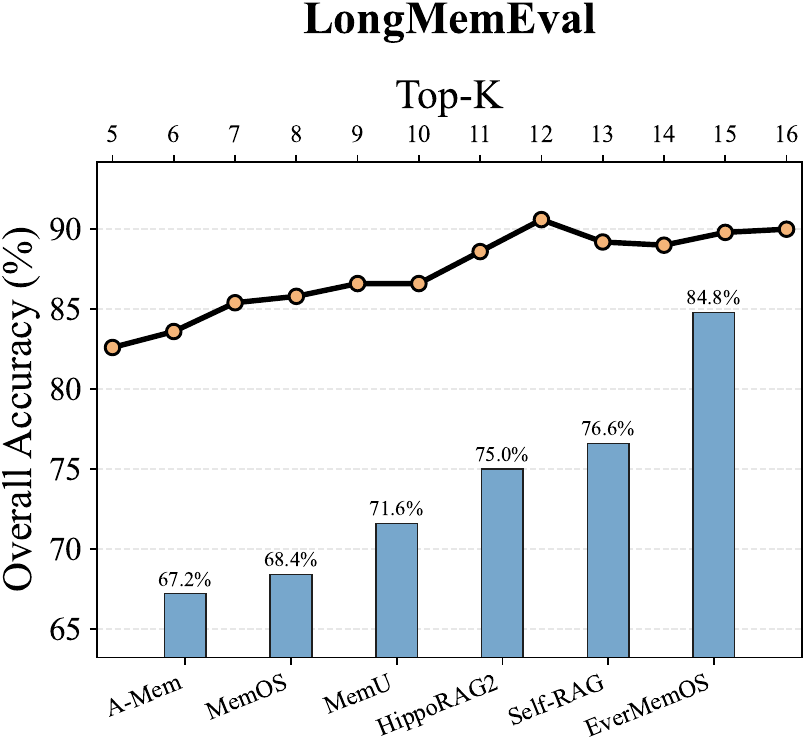}
        \label{fig:topk_sensitivity_lme}
    \end{subfigure}
    \hfill
    \begin{subfigure}[t]{0.48\linewidth}
        \centering
        \includegraphics[width=\linewidth]{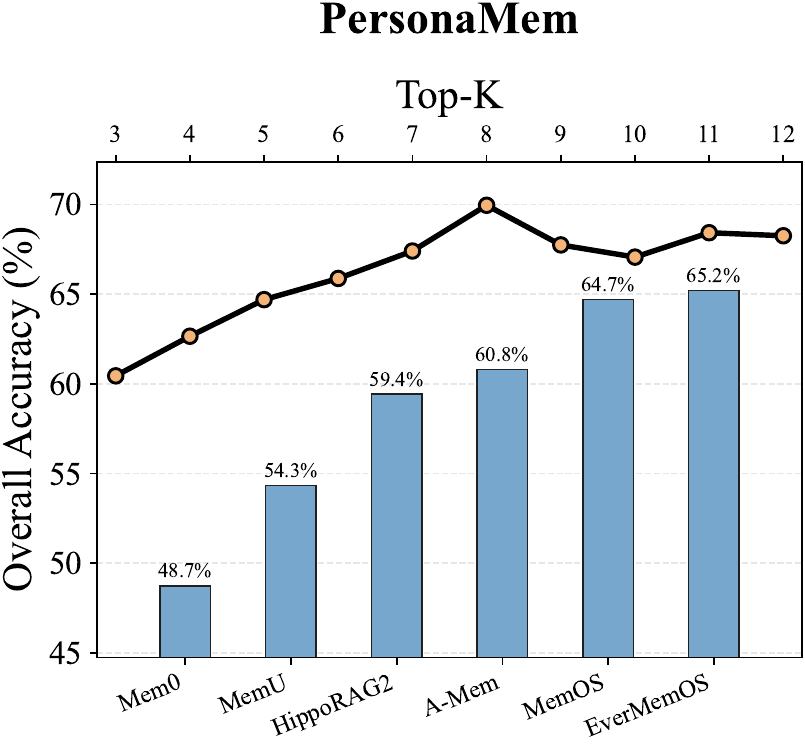}
        \label{fig:topk_sensitivity_personamem}
    \end{subfigure}
    \vspace{-1.5em}
    \caption{\textbf{Top-K sensitivity analysis on LongMemEval and PersonaMem.}}
    \label{fig:topk_sensitivity}
    \vspace{-1.8em}
\end{wrapfigure}

We analyze the effect of the main retrieval hyperparameter, the number of retrieved segments $K$, in Figure~\ref{fig:topk_sensitivity}. 
As $K$ increases, \texttt{Threader} can access more potentially relevant evidence, leading to consistent accuracy improvements in the early stage. 
However, retrieving more segments is not always beneficial: after $K$ becomes sufficiently large, additional segments introduce more irrelevant or weakly related context, and the accuracy no longer increases steadily but starts to fluctuate. 
Notably, \texttt{Threader} already surpasses all state-of-the-art baselines with a relatively small retrieval budget. 
On LongMemEval, \texttt{Threader} outperforms all baselines at Top-7, and on PersonaMem, it surpasses all baselines at Top-6. 
This suggests that \texttt{Threader} is not only effective at high retrieval budgets, but can also locate useful evidence with a compact retrieved context. 
We further analyze the retrieval hyperparameters in Eqs.~\ref{eq:dense}, \ref{eq:hybrid}, and~\ref{eq:final}
in Appendix~\ref{app:hyper}.

\subsection{Cost Analysis}\label{sec:cost}

Cost is a critical yet under-reported factor in agent memory systems. 
Prior work often reports a single aggregated number, obscuring the fundamentally different cost characteristics of memory construction and query-time serving. 
We therefore decompose system cost into two stages: 
\textbf{(1) offline memory construction}, where monetary cost (LLM calls, token usage) dominates, and 
\textbf{(2) online response}, where latency is the primary concern. 

Table~\ref{tab:cost_analysis} shows that existing methods incur substantial cost in at least one stage.
Rewrite-heavy approaches introduce significant construction overhead.
For example, processing 100K input tokens requires \textbf{21.1$\times$} and \textbf{13.0$\times$} token expansion for A-Mem and EverMemOS, respectively, along with hundreds of LLM calls and non-trivial monetary cost.
In contrast, \texttt{Threader} eliminates construction-time LLM usage and builds memory indices in 18.33s, achieving over \textbf{110$\times$} speedup compared to these methods.
It is only slightly slower than Self-RAG (18.33s vs. 13.18s), mainly due to its additional topic-coherent segmentation step before indexing.
At query time, \texttt{Threader} maintains low latency (8.66s) without retrieval-side LLM calls, incurring no extra token cost beyond final response generation.
Although Mem0 is slightly faster (8.38s) due to simpler retrieval over pre-written memory entries, \texttt{Threader} achieves comparable latency without LLM-based construction.
Despite its multi-signal design (Section~\ref{method:retrieve}), \texttt{Threader}'s retrieval remains lightweight in practice, achieving \textbf{5.9$\times$--23.0$\times$} query-time speedups over other representative baselines.
Overall, existing systems face a trade-off: LLM-based memory construction incurs high offline cost, while agentic retrieval introduces expensive query-time overhead.
\texttt{Threader} avoids this trade-off by replacing LLM-based construction with lightweight structure-aware indexing and using LLM-free retrieval, enabling both low construction overhead and efficient query-time access.

\begin{table*}[tb!]
\caption{\textbf{Cost comparison of agent memory systems across offline memory construction and online query-time serving.} MemOS is omitted due to its black-box interface, which prevents reliable cost attribution.
We report construction cost per 100K input tokens and query-time cost per query, covering runtime/latency (s), LLM calls, input and output token consumption, and monetary cost (\$). 
The best and worst reported results are highlighted in \colorbox{green!20}{green} and \colorbox{red!20}{red}, respectively.}
\label{tab:cost_analysis}

\centering
\footnotesize
\setlength{\tabcolsep}{2pt}
\renewcommand{\arraystretch}{1.05}
\providecommand{\tokcell}[2]{}
\renewcommand{\tokcell}[2]{%
    \makebox[2.8em][r]{#1}\makebox[0.9em][l]{#2}%
}
\begin{tabular}{
    @{}
    l
    |S[table-format=4.2]
     S[table-format=4.2]
     c
     c
     c
    |S[table-format=3.2]
     S[table-format=2.2]
     c
     c
     c
    @{}
}
\toprule
\multirow{2}{*}{\raisebox{-0.4em}{\textbf{Method}}}
& \multicolumn{5}{c|}{\textbf{Memory Construction}}
& \multicolumn{5}{c}{\textbf{Query-Time}} \\
\cmidrule(lr){2-6} \cmidrule(lr){7-11}
& {\textbf{Time}}
& {\textbf{\#Calls}}
& \textbf{In Tok.}
& \textbf{Out Tok.}
& \textbf{Cost \$}
& {\textbf{Latency}}
& {\textbf{\#Calls}}
& \textbf{In Tok.}
& \textbf{Out Tok.}
& \textbf{Cost \$} \\
\midrule
Self-RAG   & \cellcolor{green!20}13.18 & \cellcolor{green!20}0 & \cellcolor{green!20}0 & \cellcolor{green!20}0 & \cellcolor{green!20}0
           & 93.52 & 6.55 & \tokcell{2.3}{K} & \tokcell{0.4}{K} & 0.0016 \\
HippoRAG2  & 110.84 & 95.26 & \tokcell{249}{K} & \tokcell{38}{K} & 0.1604
           & 51.02 & 3.03 & \cellcolor{red!20}\tokcell{17.5}{K} & \tokcell{0.7}{K} & 0.0081 \\
Mem0       & 1239.62 & 47.94 & \tokcell{231}{K} & \tokcell{64}{K} & 0.1948
           & \cellcolor{green!20}8.38 & \cellcolor{green!20}0 & \cellcolor{green!20}0 & \cellcolor{green!20}0 & \cellcolor{green!20}0 \\
MemU       & \cellcolor{red!20}2448.94 & 305.65 & \tokcell{936}{K} & \tokcell{121}{K} & 0.5680
           & \cellcolor{red!20}199.47 & \cellcolor{red!20}19.09 & \tokcell{16.1}{K} & \cellcolor{red!20}\tokcell{2.5}{K} & \cellcolor{red!20}0.0112 \\
A-Mem      & 2032.45 & \cellcolor{red!20}1415.19 & \cellcolor{red!20}\tokcell{2107}{K} & \tokcell{79}{K} & \cellcolor{red!20}0.9692
           & 12.35 & 1.00 & \tokcell{9.9}{K} & \tokcell{0.9}{K} & 0.0054 \\
EverMemOS  & 2013.97 & 632.42 & \tokcell{1298}{K} & \cellcolor{red!20}\tokcell{146}{K} & 0.7528
           & 53.68 & 1.75 & \tokcell{3.1}{K} & \tokcell{0.3}{K} & 0.0017 \\
\textbf{\texttt{Threader}}        & 18.33 & \cellcolor{green!20}0 & \cellcolor{green!20}0 & \cellcolor{green!20}0 & \cellcolor{green!20}0
           & 8.66 & \cellcolor{green!20}0 & \cellcolor{green!20}0 & \cellcolor{green!20}0 & \cellcolor{green!20}0 \\
\bottomrule
\end{tabular}
\vspace{-1.5em}
\end{table*}

\section{Conclusions}

In this work, we revisit the design of memory systems for conversational agents and identify fundamental limitations of LLM-based memory construction in long-horizon, high-entropy settings. 
Through empirical analysis, we show that heavy reliance on construction-time abstraction leads to both information loss and substantial computational overhead, limiting the effectiveness and scalability of existing approaches.
We propose \texttt{Threader}, a memory framework that shifts the focus from constructing compressed representations to enabling efficient, structure-aware access over raw interactions. 
By combining incremental segmentation, multi-view representation, and multi-signal retrieval, \texttt{Threader} achieves robust and stable access to query-relevant evidence while preserving contextual coherence.
Extensive experiments demonstrate that \texttt{Threader} not only improves answer accuracy and evidence recall, but also substantially reduces both construction and query-time cost.

\subsection*{AI Use Statement}

The authors designed the algorithmic framework of \textsc{Threader} and implemented the method. Generative AI tools were used to assist in reviewing the framework and implementation, identifying potential errors, and suggesting refinements. The authors assessed these suggestions, determined which changes to incorporate, and reviewed and tested the resulting implementation.

The manuscript was drafted by the authors. Generative AI tools were used to assist with wording, organization, and technical presentation, and to improve grammar, clarity, and readability. The authors reviewed and revised the AI-assisted edits to ensure consistency with their intended meaning, the implemented method, and the reported experimental results.

As part of the experimental pipeline, GPT-4.1-mini was used to annotate fine-grained intra-session topic boundaries for segmenter training and evaluation. GPT-4.1-mini was also used as an LLM-as-a-Judge to evaluate answer correctness on LongMemEval by comparing generated responses against reference answers under the specified grading protocol. These uses are described in the corresponding annotation and evaluation sections.

The research premises and hypotheses were established by the authors. The authors made all final scientific decisions regarding the method, experimental design, interpretation of results, and conclusions. The authors take full responsibility for the final content of this work, including its methods, analyses, results, claims, and AI-assisted materials.

\subsection*{Ethics statement}

We strictly follow the data-usage and security requirements when interacting with the OpenAI API and Alibaba Qwen API services. This study uses only publicly available datasets and does not involve collecting new personal data from participants. To mitigate potential information leakage, we restrict API inputs to benchmark content, remove identifying information where applicable, and enable provider-side privacy controls (e.g., disabling data sharing or retention for model improvement when available). All API credentials are securely managed, and model outputs are used solely for academic research and evaluation purposes.

\subsection*{Reproducibility statement}

Code is available at \url{https://github.com/Donghua-Cai/Threader}, as linked in the abstract. The paper describes the algorithmic framework in Section~\ref{method} and provides pseudocode in Appendix~\ref{algorithm}. Implementation details, segmenter training procedures, and retrieval hyperparameters are documented in Appendices~\ref{app:implement}, \ref{app:segmenter_training}, and \ref{app:hyper}, respectively. We also provide the LLM-as-a-Judge protocol and prompt in Appendix~\ref{app:judge} and the response-generation prompts in Appendix~\ref{app:response prompt}. These materials specify the main components and experimental settings needed to reproduce our method and evaluation.

\bibliography{iclr2027_conference}
\bibliographystyle{iclr2027_conference}

\appendix

\section{Limitations and Future Work}

\subsection{Limitations}
\label{app:limitation}

While \texttt{Threader} demonstrates strong performance and efficiency in long-horizon conversational settings, we identify several potential limitations.

\textbf{Dependence on segmentation quality.}
Our approach relies on topic-coherent segmentation to organize raw interactions into retrieval units. 
Although the proposed segmenter is lightweight and effective in practice, segmentation errors may propagate to downstream retrieval. 
In particular, if semantically related evidence is split across segment boundaries, it may reduce recall. 
Designing more robust segmentation strategies or incorporating boundary-aware retrieval remains an important direction for future work.

\textbf{Scalability under extremely large memory.}
While our indexing and retrieval pipeline is efficient, scaling to extremely large or lifelong interaction histories may introduce additional storage and retrieval challenges. 
Incorporating memory pruning, compression, or hierarchical indexing mechanisms could further improve scalability.

\textbf{Limited abstraction of long-range dependencies.}
\texttt{Threader} emphasizes preserving fine-grained evidence and deferring abstraction to query time. 
However, some tasks may benefit from higher-level abstraction or long-range reasoning that integrates information across distant segments. 
Extending the framework to support adaptive abstraction—combining raw evidence with learned summaries—could improve performance in such scenarios.

\subsection{Future Work} In future work, we will extend \texttt{Threader} along two main directions. 
First, we will broaden the current factual-memory framework toward more general agent memory scenarios, especially experiential memory for action-oriented agents. 
Beyond preserving and retrieving user-provided facts in long-horizon dialogue, future extensions will consider task trajectories, action outcomes, tool-use traces, failure cases, and reusable execution strategies as memory units. 
This requires adapting \texttt{Threader}'s topic-coherent segmentation, multi-view representation, and multi-signal retrieval to heterogeneous agent experiences that involve observations, actions, tool outputs, and environmental feedback. 
Such an extension would allow \texttt{Threader} to support not only evidence recall in personalized dialogue, but also long-horizon planning and action execution in tool-use, embodied, web, and coding agents. 
Second, we will work toward stronger memory governance for privacy protection, user control, and responsible memory management. 
Important directions include user-facing mechanisms for memory inspection, editing, deletion, and selective forgetting, as well as privacy-preserving storage, access control, retention policies, and safeguards against unauthorized profiling or inappropriate memory use. 
These extensions aim to make \texttt{Threader} more general, controllable, and trustworthy for real-world long-term agent deployments.

\section{Additional Related Works}
\subsection{Experiential Memory in Agent Learning}

Experiential memory aims to improve an agent's future task-solving ability by storing and reusing past interaction experience, such as trajectories, success/failure cases, skills, and high-level strategies. 
Early reflective agents use verbal feedback from failed attempts to improve subsequent trials, enabling within-task self-improvement through accumulated reflections \citep{shinn2023reflexion}. 
Beyond single-task retry, ExpeL stores past trajectories and extracts cross-task insights from successful and failed experiences, allowing agents to retrieve similar experiences or follow distilled strategies at inference time \citep{zhao2024expel}. 
Other systems further transform experience into executable skills, workflows, or reusable policies for open-ended environments and tool-use tasks \citep{wang2023voyager, wang2024agent, ouyang2025reasoningbank}. 
These methods are effective for action-oriented agents, where memory primarily supports planning, exploration, and policy improvement. 
In contrast, our work focuses on factual memory in long-horizon human--assistant dialogue, where the key challenge is not to learn reusable action strategies, but to preserve and retrieve fine-grained user-specific evidence under long, entangled interaction histories.

\subsection{Working Memory in Task Execution}

Working memory maintains transient task state and intermediate context during ongoing reasoning or execution. 
Unlike experiential memory, which accumulates reusable knowledge across episodes, working memory is usually invoked within a single task trajectory to manage limited context budgets, preserve important intermediate information, and prevent goal drift. 
Recent work explores memory mechanisms for long-horizon task execution, including summarizing or folding interaction traces into compact state representations, selectively retaining task-relevant evidence, and organizing subgoals or intermediate artifacts for later reasoning steps \citep{yu2025memagent, wu2025resum, hu2025hiagent}. 
Such mechanisms are particularly useful for web navigation, deep research, tool-use, and multi-step reasoning, where the agent must continuously update its internal workspace as observations and actions accumulate \citep{sun2025scaling, ye2025agentfold, li2026deepagent}. 
However, working memory is typically optimized for maintaining task progress within an active execution process, rather than building a persistent, queryable record of long-term user history. 
Our setting instead requires a durable memory bank that supports future factual recall across sessions, making evidence preservation and retrieval fidelity central design concerns.

\subsection{Parametric and Latent Memory.}
Agent memory can also be realized in parametric or latent forms, beyond explicit token-level storage~\citep{hu2025memory}. 
Parametric memory stores information in learnable parameters, either by editing the backbone or by attaching additional parameter modules. 
Representative model-editing methods, such as MEND~\citep{mitchell2021fast} and ROME~\citep{meng2022locating}, inject factual knowledge through localized parameter updates, while adapter-style approaches such as K-Adapter~\citep{wang2021k} store additional knowledge in modular parameters. 
Latent memory instead stores information in continuous internal representations, such as hidden states, memory tokens, KV caches, or compressed embeddings. 
For example, MemoryLLM and its extension M+ introduce persistent memory tokens or cross-layer token pools in the latent space of the transformer for self-updatable long-term knowledge storage~\citep{wang2024memoryllm,wang2025m+}, while AutoCompressor compresses long contexts into compact summary vectors for later reuse~\citep{chevalier2023adapting}. 

Although parametric and latent memory can compactly internalize knowledge and reduce explicit retrieval overhead, they are less aligned with our target setting of long-term personalized dialogue. 
Parametric updates are often model-dependent, costly to revise, and vulnerable to interference or catastrophic forgetting, while latent memory is difficult to inspect, edit, or attribute to concrete dialogue evidence. 
\texttt{Threader} therefore adopts token-level factual memory: instead of internalizing dialogue history into parameters or hidden states, it preserves raw dialogue content as externally accessible topic-coherent evidence segments, prioritizing transparency, controllability, and evidence fidelity.

\begin{table*}[!b]
\centering
\small
\setlength{\tabcolsep}{6pt}
\renewcommand{\arraystretch}{1.08}
\caption{\textbf{Dataset statistics.}
We report the number of QA pairs, the number of samples, and dialogue-level statistics for each benchmark.
All metrics with ``/'' denote average values, including sessions per sample, tokens per session, turns per session, and tokens per turn.}
\label{tab:dataset_statistics}
\resizebox{\textwidth}{!}{
\begin{tabular}{lcccccc}
\toprule
\textbf{Dataset}
& \textbf{\# QA}
& \textbf{\# Sample}
& \textbf{Sess./Sample}
& \textbf{Tok./Sess.}
& \textbf{Turn/Sess.}
& \textbf{Tok./Turn} \\
\midrule
LongMemEval & 500 & 500 & 47.7 & 2,160.73 & 5.12 & 421.88 \\
PersonaMem  & 589 & 37 & 1 & 26,165.38 & 174.65 & 149.82 \\
\bottomrule
\end{tabular}
}
\end{table*}

\section{Algorithm}
\label{algorithm}

The overall procedure of \texttt{Threader} is summarized in Algorithm~\ref{alg:vsm}.

\section{Datasets}
\label{app:datasets}

We provide detailed descriptions of the two long-term dialogue memory benchmarks used in our evaluation: LongMemEval \citep{wu2024longmemeval} and PersonaMem \citep{jiang2025know}.
Both datasets are designed to evaluate whether an agent can retrieve and use information from long interaction histories, but they emphasize different aspects of memory.
LongMemEval focuses on multi-session factual and temporal memory over user--assistant conversations, while PersonaMem stresses long-session personalization, preference understanding, and user-specific response generation.
The statistics of the two datasets are summarized in Table~\ref{tab:dataset_statistics}.

\paragraph{LongMemEval.}
LongMemEval is a long-term memory benchmark for evaluating whether dialogue agents can answer questions grounded in previous user--assistant interactions.
Each test instance consists of a user query, a long multi-session conversation history, and a gold answer.
The model is required to identify the relevant evidence from the historical dialogue and generate an answer consistent with the gold response.
Compared with standard open-domain QA, LongMemEval places stronger emphasis on memory retrieval, since the information needed to answer the question is usually contained in past conversations rather than in the model's parametric knowledge.

LongMemEval contains 500 QA instances, with an average session length of approximately 2,160 tokens.
Its questions cover several question types that assess different long-term memory abilities, including single-session user information recall, single-session assistant information recall, preference recall, multi-session reasoning, knowledge update, and temporal reasoning.
These question types make LongMemEval suitable for evaluating whether a memory system can locate sparse evidence, distinguish outdated and updated information, and reason over events mentioned across different sessions.

\begin{algorithm}[H]
\caption{\texttt{Threader}}
\label{alg:vsm}
\footnotesize

\KwInput{
$C=\{S_1,\dots,S_n\}$: conversation history; 
$S_i=(t_1,\dots,t_m)$: the $i$-th session; 
$t_j=(u_j,a_j)$: the $j$-th user-assistant turn; 
$q$: the user query; 
$g_{\theta}$: the incremental topic-coherent segmenter; 
$f_{\mathrm{enc}}$: the encoder; 
$\mathrm{BM25}(\cdot,\cdot)$: the lexical retriever; 
$\mathrm{Rerank}(\cdot,\cdot)$: the cross-encoder reranker.
}

\KwOutput{
The final generated response for the given user query $q$.
}

\Comment{Incremental Topic-Coherent Segmentation}

Construct topic-coherent evidence segments from the conversation history $C$\;

\ForEach{session $S_i=(t_1,\dots,t_m)$ in $C$}{
    Initialize the current segment $s'\leftarrow(t_1)$ and the segment set $\mathcal{S}_i\leftarrow\emptyset$\;
    
    \ForEach{incoming turn $t_{j+1}$ in $S_i$}{
        Construct the segmentation input by concatenating the current segment $s'$ and the incoming turn $t_{j+1}$\;
        
        Predict whether $t_{j+1}$ should continue the current segment or start a new segment using $g_{\theta}$\;
        
        \eIf{the prediction is \textsc{CONTINUE}}{
            Append $t_{j+1}$ to the current segment $s'$\;
        }{
            Add $s'$ to $\mathcal{S}_i$ and start a new segment with $t_{j+1}$\;
        }
    }
    
    Add the final segment $s'$ to $\mathcal{S}_i$\;
}

Obtain the full segment set $\mathcal{S}=\bigcup_i\mathcal{S}_i$\;

\Comment{Multi-View Segment Representation}

\ForEach{topic-coherent segment $s_i=(t_{i,1},\dots,t_{i,m})$ in $\mathcal{S}$, where $t_{i,j}=(u_{i,j},a_{i,j})$}{
    Construct three complementary views:
    $x_i^{(u)}=\mathrm{Concat}(u_{i,1},\dots,u_{i,m})$,
    $x_i^{(a)}=\mathrm{Concat}(a_{i,1},\dots,a_{i,m})$,
    and $x_i^{(f)}=\mathrm{Concat}(u_{i,1},a_{i,1},\dots,u_{i,m},a_{i,m})$\;
    
    Encode each view using the shared encoder:
    $e_i^{(v)}=f_{\mathrm{enc}}(x_i^{(v)}),\ v\in\{u,a,f\}$\;
    
    Store $\{e_i^{(u)},e_i^{(a)},e_i^{(f)}\}$ for dense retrieval\;
}

Build dense indices over the multi-view segment representations and a BM25 index over raw segments\;

\Comment{Stage I: High-Recall Segment Retrieval}

Encode the user query $q$ using the shared encoder $f_{\mathrm{enc}}$:$z(q) = f_{\mathrm{enc}}(q)$ \;

Compute the multi-view dense semantic score for each segment via Eq.~\eqref{eq:dense}\;

Compute the lexical matching score for each segment\;

Construct the high-recall candidate set $\mathcal{C}$ by combining dense and lexical retrieval results via Eq.~\eqref{eq:pool}\;

\Comment{Stage II: Evidence-Level Refinement}

\ForEach{candidate segment $s_i$ in $\mathcal{C}$}{
    Compute message-level similarities between $q$ and each user/assistant message in $s_i$\;
    
    Select the top-$K_2$ message-level similarities and average them to obtain the refined dense score via Eq.~\eqref{eq:msg}\;
    
    Combine the refined dense signal with the lexical matching signal via Eq.~\eqref{eq:hybrid}\;
    
    Apply the cross-encoder reranker and aggregate it with the hybrid score to obtain the final segment score via Eq.~\eqref{eq:final}\;
}

Select the top-$K$ segments from $\mathcal{C}$ according to the final segment score\;

Obtain the retrieved evidence segments $\mathcal{R}$\;

\Comment{Response Generation}

Generate the final response by conditioning the response model on the user query $q$\;

\end{algorithm}

We briefly describe the question types in LongMemEval as follows.

\begin{itemize}[leftmargin=2em,itemsep=0.25em,topsep=0.25em]
    \item \textbf{Single-session user information recall} evaluates whether the model can recall facts explicitly mentioned by the user within a past session, such as personal attributes, plans, constraints, or experiences.
    
    \item \textbf{Single-session assistant information recall} evaluates whether the model can remember information provided by the assistant in previous interactions, such as recommendations, explanations, or decisions made during the dialogue.
    
    \item \textbf{Single-session preference recall} evaluates whether the model can identify user preferences expressed in prior conversations, including likes, dislikes, habits, or requirements.
    
    \item \textbf{Multi-session reasoning} requires the model to combine evidence distributed across multiple sessions. This question type tests whether the memory system can retrieve complementary pieces of information rather than relying on a single isolated turn.
    
    \item \textbf{Knowledge update} evaluates whether the model can handle evolving information. The model must identify the most recent or valid state when earlier information is later revised, corrected, or superseded.
    
    \item \textbf{Temporal reasoning} requires the model to reason about time-sensitive information, such as event order, relative dates, or changes across sessions.
\end{itemize}

\paragraph{PersonaMem.}
PersonaMem is a long-session personalization benchmark designed to evaluate whether a dialogue agent can understand and use user-specific information over extended interactions.
Each instance contains a long user--assistant dialogue history and a question that requires the model to recall, infer, or apply information about the user.
Compared with LongMemEval, PersonaMem contains substantially longer sessions, with an average session length of approximately 26,165 tokens, making it especially suitable for testing memory systems under long-context and high-entropy interaction settings.

PersonaMem contains 589 QA instances and covers seven question types.
These question types evaluate not only direct fact recall, but also the model's ability to generalize user preferences, explain preference changes, and make personalized suggestions.
This makes PersonaMem a challenging benchmark for long-term user modeling, since relevant evidence is often sparse, interleaved with unrelated topics, and distributed across a long interaction history.

We describe the seven question types in PersonaMem as follows.

\begin{itemize}[leftmargin=2em,itemsep=0.25em,topsep=0.25em]
    \item \textbf{Generalizing to new scenarios} evaluates whether the model can apply previously observed user preferences or behaviors to a new situation. The answer is not always a direct copy of a past utterance, but should be consistent with the user's established profile.
    
    \item \textbf{Preference-aligned recommendation} requires the model to provide recommendations that match the user's historical preferences, constraints, or goals. This question type evaluates whether the model can retrieve and synthesize relevant preference evidence.
    
    \item \textbf{Recalling user-shared facts} evaluates whether the model can recall factual information explicitly shared by the user, such as background, interests, routines, or personal constraints.
    
    \item \textbf{Recalling facts mentioned by the user} focuses on information that appeared in the user's prior messages, including facts about entities, events, or external situations discussed by the user.
    
    \item \textbf{Recalling reasons behind previous updates} evaluates whether the model can remember why a preference, plan, or decision changed. This requires recovering not only the updated state, but also the explanatory evidence behind the update.
    
    \item \textbf{Suggesting new ideas} requires the model to propose new options or actions based on the user's long-term preferences and past interactions. This question type tests whether memory can support personalized generation beyond direct extraction.
    
    \item \textbf{Tracking full preference evolution} evaluates whether the model can follow how a user's preference changes over time. The model must distinguish earlier preferences from later updates and generate an answer consistent with the complete preference trajectory.
\end{itemize}

Overall, the two datasets provide complementary evaluation settings.
LongMemEval offers a controlled multi-session memory QA benchmark with diverse question types, while PersonaMem presents longer and more personalized interaction histories that better stress-test memory systems under realistic long-session conditions.
Together, they allow us to evaluate both factual memory retrieval and personalized memory utilization in long-horizon user--assistant dialogue.

Although LoCoMo \citep{maharana2024evaluating} has been widely adopted in prior work for evaluating long-term dialogue memory, we do not include it in our main evaluation because it differs from our target setting in both session length and interaction format.
First, each LoCoMo session contains only about 586 tokens on average, making it substantially shorter than the long-session histories considered in this work.
Second, LoCoMo is constructed from simulated two-person conversations, whereas our focus is long-term personalized memory for user--assistant interaction histories.
We therefore use LongMemEval and PersonaMem as the primary benchmarks, as they better align with the long-horizon human--assistant setting studied in this paper.

\section{Baseline and Implementation Details}
\label{app:implement}
Our baselines cover both persistent memory systems and adaptive retrieval frameworks. 
For methods with official public implementations, we follow the released codebases and recommended default settings whenever they are compatible with our evaluation pipeline. 
In particular, several baselines use OpenAI embedding models, such as \texttt{text-embedding-3-large} and \texttt{text-embedding-ada-002}, as their default embedding backbones; we keep these default configurations and follow the official OpenAI embedding API documentation.\footnote{\url{https://platform.openai.com/docs/api-reference/embeddings}} 
For Self-RAG~\citep{asai2024selfrag}, since the original release is a specially fine-tuned model, we instead use a LangGraph-based re-implementation of its retrieve-generate-reflect workflow with our shared backbone models, to ensure a fair comparison across methods. All public datasets, pretrained models, and baseline codebases used in this work are credited to their original authors and used under their respective licenses and terms of use.

\paragraph{Self-RAG.}
Self-RAG is not a persistent memory system in the strict sense, but it serves as a strong adaptive retrieval baseline \cite{asai2024selfrag}. It trains the model to decide when retrieval is necessary during generation and to emit reflection tokens that critique retrieval relevance, evidential support, and response utility, thereby forming a retrieve-generate-critique loop. Compared with standard RAG pipelines that retrieve in a fixed manner, Self-RAG improves factual grounding while preserving the general-purpose flexibility of the underlying language model. In our implementation, we use a LangGraph-based re-implementation with GPT-4.1-mini as the general backbone LLM, and instantiate the dense retriever with the \texttt{facebook/contriever-msmarco} \citep{izacard2021unsupervised}. 
The retriever encodes both queries and dialogue-history chunks for similarity-based retrieval, and the retrieved chunks are then fed into our unified response model for answer generation.

\paragraph{HippoRAG2.}
We use HippoRAG2, introduced in \textit{From RAG to Memory}, as a structure-aware retrieval baseline for non-parametric continual learning \citep{gutierrez2025rag}. The method builds an open knowledge graph over the corpus, performs deeper passage integration, and applies query-conditioned \textit{Personalized PageRank} to capture multi-hop associations and relational propagation beyond plain semantic matching. By combining dense retrieval signals, graph-based propagation, and more effective online use of an LLM, HippoRAG2 moves beyond standard vector RAG toward a more organized and continually extensible external memory mechanism. In our implementation, we use the official HippoRAG2 framework with GPT-4.1-mini as the general construction LLM and \texttt{paraphrase-multilingual-MiniLM-226
L12-v2} \citep{reimers-2019-sentence-bert} as embedding model. 
The retrieved memory chunks are then fed into our unified response model for answer generation.

\paragraph{Mem0.}
Mem0 models long-term memory as a module decoupled from response generation and maintains an external memory store through incremental extraction, matching, and update operations \citep{chhikara2025mem0}. 
At each step, it extracts salient facts from the current interaction, matches them against existing memories, and uses an LLM-based update policy to insert, merge, delete, or discard candidate memories. 
In our implementation, we use the official Mem0 framework with GPT-4.1-mini as the general memory construction LLM and keep its default embedding model, \texttt{text-embedding-3-large}.
The retrieved memories are then fed into the shared response LLM specified in Section~\ref{sec:setup} for answer generation.

\paragraph{MemU.}
MemU is an always-on proactive memory framework designed for long-running AI agents.\footnote{\url{https://github.com/NevaMind-AI/memU}}  It organizes memory through a file-system-like hierarchy of \textit{Resources}, \textit{Items}, and \textit{Categories}, where raw inputs are preserved as resources, extracted facts and preferences are stored as items, and topic-level categories support higher-level context assembly. Its \texttt{memorize} pipeline continuously ingests interactions, extracts memory items, updates categories, and cross-references existing memories. During retrieval, MemU supports both embedding-based RAG for fast context loading and LLM-based retrieval for deeper intent prediction and query refinement. This design enables agents to maintain structured memory, anticipate user needs, and surface relevant context proactively in long-running settings. In our implementation, we use the official MemU framework with GPT-4.1-mini as the general memory construction LLM and keep its default embedding model, \texttt{text-embedding-ada-002}.
The retrieved memories are then fed into the shared response LLM specified in Section~\ref{sec:setup} for answer generation.

\paragraph{A-Mem.}
A-Mem formulates memory management as an agentic note-taking process inspired by the Zettelkasten paradigm \cite{xu2025mem}. Each interaction is converted into a structured note containing content, timestamps, keywords, tags, contextual descriptions, and embeddings. The system then retrieves related notes and lets an LLM controller decide how to revise existing notes, create links, and evolve the memory graph. Compared with conventional memory stores that mainly support storage and retrieval, A-Mem places greater emphasis on memory organization and self-evolution. In our implementation, we use the official A-Mem framework with GPT-4.1-mini as the general memory construction LLM and keep its default embedding model, \texttt{all-MiniLM-L6-v2} \citep{reimers-2019-sentence-bert}.
The retrieved memories are then fed into the shared response LLM specified in Section~\ref{sec:setup} for answer generation.

\paragraph{MemOS.}
MemOS treats memory as a first-class system resource and introduces the unified \textit{MemCube} abstraction over plaintext, activation-based, and parameter-level memory \citep{li2025memos}. 
Rather than reducing memory to a single vector store, it coordinates memory writing, maintenance, loading, and retrieval through layered scheduling and lifecycle management mechanisms. 
This OS-style design is intended to support long-horizon reasoning, continual personalization, and reusable knowledge across tasks and environments within a single memory framework. 
In our implementation, we use the official MemOS framework and set GPT-4.1-mini as the general model. 
Since MemOS provides a high-level system interface with limited access to its internal scheduling details, we keep its internal memory scheduling and retrieval configuration unchanged and only standardize the external LLM backbone for fair comparison.

\paragraph{EverMemOS.}
EverMemOS extends the operating-system view of memory by emphasizing self-organization for structured long-horizon reasoning \cite{hu2026evermemos}. It first converts interaction streams into structured \textit{MemCells} that capture episodic traces, atomic facts, and short-term foresight signals, and then consolidates them into higher-level \textit{MemScenes} that encode themes, user profiles, and stable semantic structure. During inference, EverMemOS performs reconstructive recollection, assembling task-relevant context from structured long-term memory rather than simply retrieving a few isolated fragments. This design makes it particularly suitable for scenarios where memory must remain coherent across extended interactions. In our implementation, we use the official EverMemOS framework with GPT-4.1-mini as the general memory construction LLM and keep its default embedding model, \texttt{Qwen3-Embedding-4B} \citep{zhang2025qwen3} and reranking model, \texttt{Qwen3-Reranker-4B} \citep{zhang2025qwen3}.
The retrieved memories are then fed into the shared response LLM specified in Section~\ref{sec:setup} for answer generation.

\paragraph{\texttt{Threader}.}
\texttt{Threader} follows a lightweight, non-LLM memory construction design.
Specifically, it uses \texttt{paraphrase-multilingual-MiniLM-L12-v2}~\citep{reimers-2019-sentence-bert}
as the embedding encoder and \texttt{bge-reranker-base}~\citep{bge_embedding}
as the cross-encoder reranker.
Both models are lightweight off-the-shelf encoders and do not require LLM calls during memory construction.
Inputs exceeding the encoder context window are handled by the model's default truncation behavior.
For longer real-world text units, \texttt{Threader} can be instantiated with longer-context embedding or reranking models without changing the framework.

\section{Additional Experimental Results and Analysis}
\subsection{Detailed Memory Fidelity Analysis of Mem0 and MemU}
\label{app:memory analysis}

\begin{table}[!tb]
\centering
\small
\setlength{\tabcolsep}{8pt}
\renewcommand{\arraystretch}{1.12}
\caption{\textbf{Statistics of the constructed motivation data.}
We progressively concatenate adjacent sessions in LongMemEval to create longer-session settings.
The number of QA examples and the total dialogue content are kept unchanged, while the average number of sessions per sample decreases and the average session length increases.}
\label{tab:motivation_data_statistics}
\begin{tabular}{lccc}
\toprule
\textbf{Name} & \textbf{Setting} & \textbf{Sessions / Sample} & \textbf{Tokens / Session} \\
\midrule
LME & Original & 47.65 & 2,311.21 \\
LME-M2 & 2 session concatenation & 23.55 & 4,676.53 \\
LME-M4 & 4 session concatenation & 11.50 & 9,576.78 \\
LME-M8 & 8 session concatenation & 5.48  & 20,085.12 \\
\bottomrule
\end{tabular}
\end{table}

We construct the motivation dataset from LongMemEval by sampling 60 instances, with 10 instances from each question type. For each sample, we keep the original QA pairs, evidence annotations, and total dialogue content unchanged, and only modify the session granularity by concatenating adjacent sessions. This yields four settings: the original data, 2$\times$, 4$\times$, and 8$\times$ session concatenation. As shown in Table~\ref{tab:motivation_data_statistics}, this process progressively reduces the average number of sessions per sample from 47.65 to 5.48, while increasing the average session length from 2,311.21 to 20,085.12 tokens. We then evaluate two representative rewrite-based memory methods, Mem0 and MemU, on these motivation datasets to examine how LLM-based memory construction behaves as individual sessions become longer and more information-dense.
Both methods perform memory construction at the session level rather than the turn level, making them directly affected by changes in session granularity.

\begin{figure*}[!tb]
\centering

\begin{subfigure}[t]{0.4\textwidth}
    \centering
    \includegraphics[width=\linewidth]{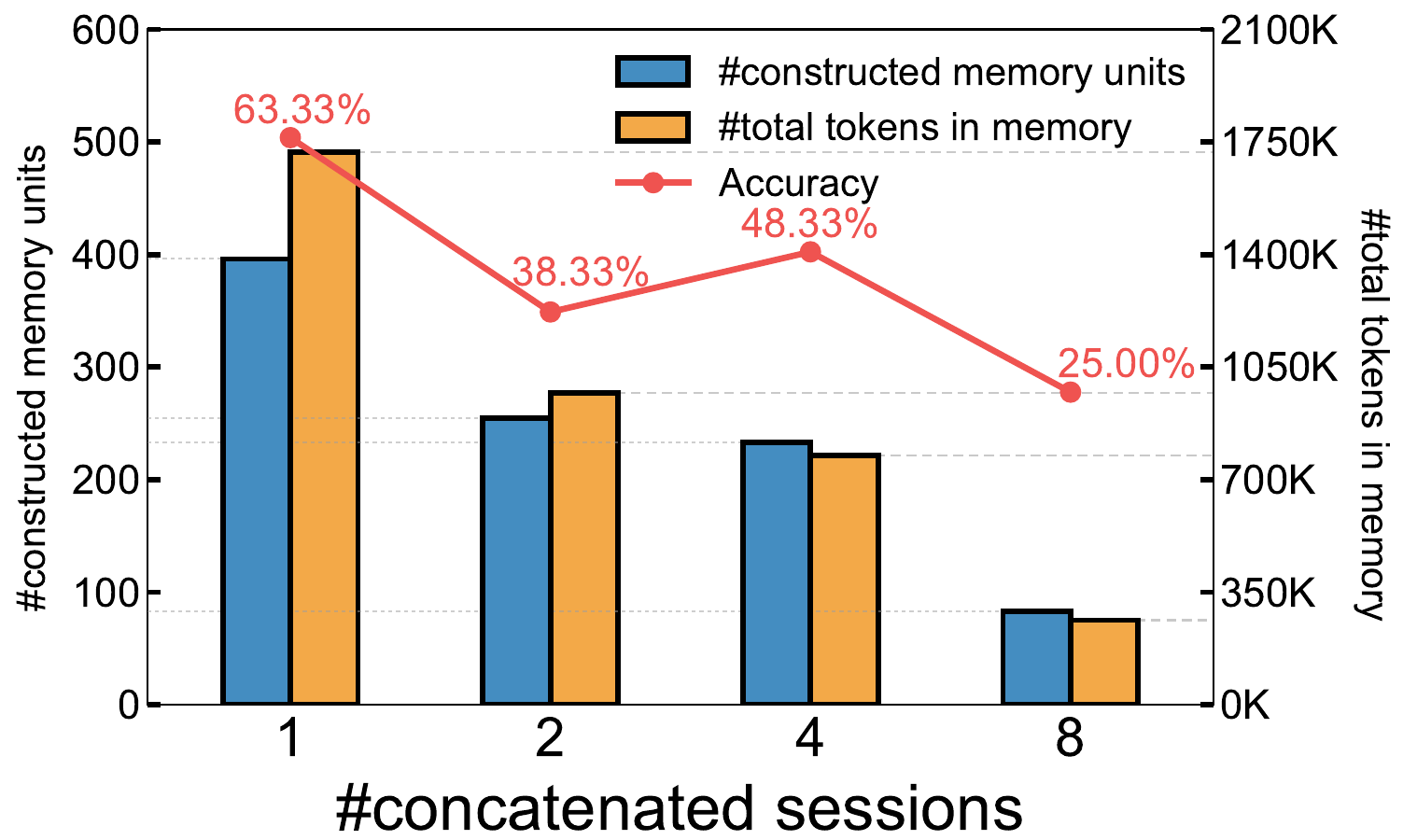}
    \captionsetup{font=footnotesize}
    \caption{Accuracy and memory size of MemU.}
    \label{fig:memu_memory}
\end{subfigure}\hspace{0.08\linewidth}
\begin{subfigure}[t]{0.4\textwidth}
    \centering
    \includegraphics[width=\linewidth]{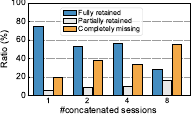}
    \captionsetup{font=footnotesize}
    \caption{Evidence preservation in MemU.}
    \label{fig:memu_coverage}
\end{subfigure}\hfill
\caption{Limitations of existing LLM-based memory construction methods (MemU). As interaction length
and information entanglement increase (simulated via session concatenation with LongMemEval),
MemU exhibits (a) degraded QA accuracy and shrinking memory size, (b) increasing
loss of query-relevant evidence.}
\label{fig:four_panel}
\end{figure*}

The main results are summarized in Section~\ref{sec:intro}. 
As shown in Fig.~\ref{fig:mem0_memory} and Fig.~\ref{fig:memu_memory}, when sessions become longer after concatenation, both Mem0 and MemU construct fewer and less informative memory entries, indicating that the memory writing process becomes increasingly lossy. 
Since these methods construct memory by directly rewriting each session into memory entries, longer sessions force the LLM-based memory writer to process more heterogeneous information within a single input. 
This makes query-relevant facts, especially those located in the middle of long sessions or surrounded by unrelated topics, more likely to be compressed away or omitted, consistent with the well-known lost-in-the-middle behavior of long-context LLMs~\citep{liu2024lost}.
This degradation further affects downstream QA accuracy, which drops as the session length increases. 
We also observe that the degradation is not smooth: QA correctness fluctuates across different concatenation settings, suggesting that rewrite-based memory construction is unstable under long and high-entropy sessions.

To diagnose this issue, we further examine whether the annotated evidence is preserved in the constructed memory banks. 
Specifically, we manually inspect the memory entries produced by Mem0 and MemU for each QA instance and compare them against the annotated evidence required to answer the question. 
If all required evidence is fully preserved in the memory bank, we label the instance as \textit{Fully Retained}. 
If none of the required evidence is preserved, we label it as \textit{Completely Missing}. 
For QA instances that require multiple pieces of evidence, if only a subset of the required evidence is retained, we label the instance as \textit{Partially Retained}. 
Fig.~\ref{fig:mem0_coverage} and Fig.~\ref{fig:memu_coverage} show that a substantial portion of query-relevant evidence is only partially retained or completely missing, confirming that many failures originate from the memory construction stage rather than the final response generation alone.

Based on the above results, we identify two characteristic failure patterns of rewrite-based memory construction under increasingly long sessions.

\textbf{Phenomenon 1: Memory degradation.}
As individual sessions become longer and more information-dense, the constructed memory becomes progressively smaller and less informative. In other words, the memory system stores fewer memory items and fewer words, even though the original dialogue contains the same amount of information. This indicates that the memory writer increasingly compresses, filters, or omits details when processing long and heterogeneous sessions, causing query-relevant evidence to disappear from the memory bank.

\textbf{Phenomenon 2: Uncontrolled memory retention.}
Memory retention is not monotonic or controllable across difficulty settings. Evidence retained in a harder setting is not necessarily retained in an easier setting, and vice versa. This suggests that rewrite-based memory construction is sensitive to session boundaries and local context organization: whether a specific fact is preserved depends not only on its importance, but also on how it is surrounded by other information during memory writing.

\begin{figure}[!tb]
    \centering
    \includegraphics[width=\linewidth]{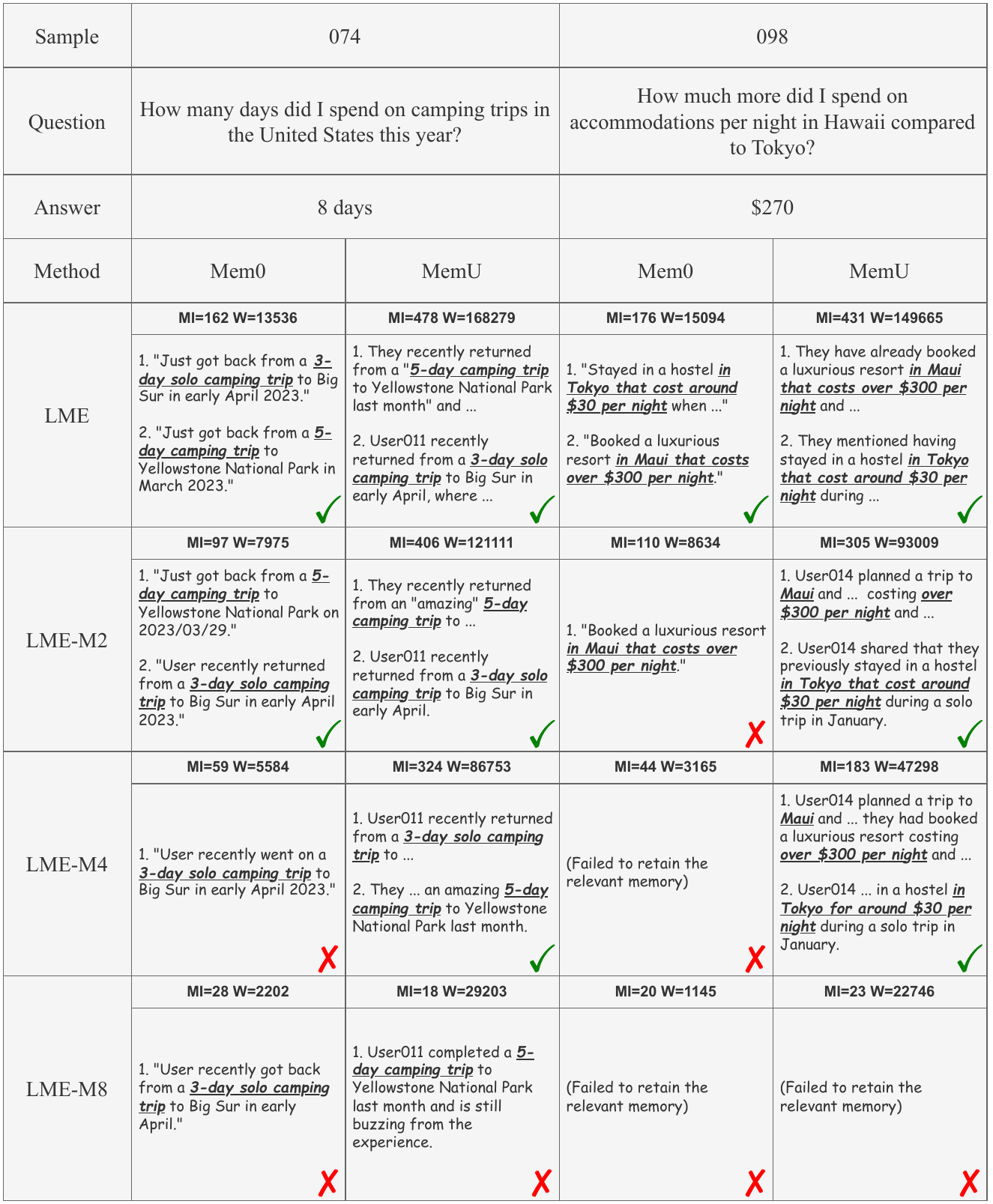}
    \caption{\textbf{Case study of memory degradation.}
As the concatenation factor increases, Mem0 and MemU construct fewer memory items and fewer words, and progressively lose query-relevant evidence. MI denotes the number of memory items, and W denotes the total number of words in the constructed memory.}
\label{fig:app_memory_degradation}
\end{figure}

\paragraph{Case Study.}
We provide two case studies to illustrate the above phenomena. Figure~\ref{fig:app_memory_degradation} illustrates \textbf{Phenomenon 1}. In Sample 074 and Sample 098, as the session concatenation factor increases from the original LongMemEval setting to the 8$\times$ setting, both Mem0 and MemU retain fewer memory items and fewer words. More importantly, the retained memories gradually lose the evidence required for answering the query. For example, in Sample 074, the answer requires aggregating two camping trips, i.e., a 3-day trip to Big Sur and a 5-day trip to Yellowstone. However, under longer-session settings, the constructed memory increasingly misses one of these two pieces of evidence, making it impossible to recover the correct answer. Similarly, in Sample 098, the answer requires comparing the accommodation costs in Maui and Tokyo, but the relevant evidence becomes partially retained or completely missing as sessions become longer. Here, MI denotes the number of constructed memory items, and W denotes the total number of words in the constructed memory. The consistent reduction of MI and W shows that long-session construction causes memory degradation rather than merely retrieval failure. 

\begin{figure}[!tb]
    \centering
    \includegraphics[width=\linewidth]{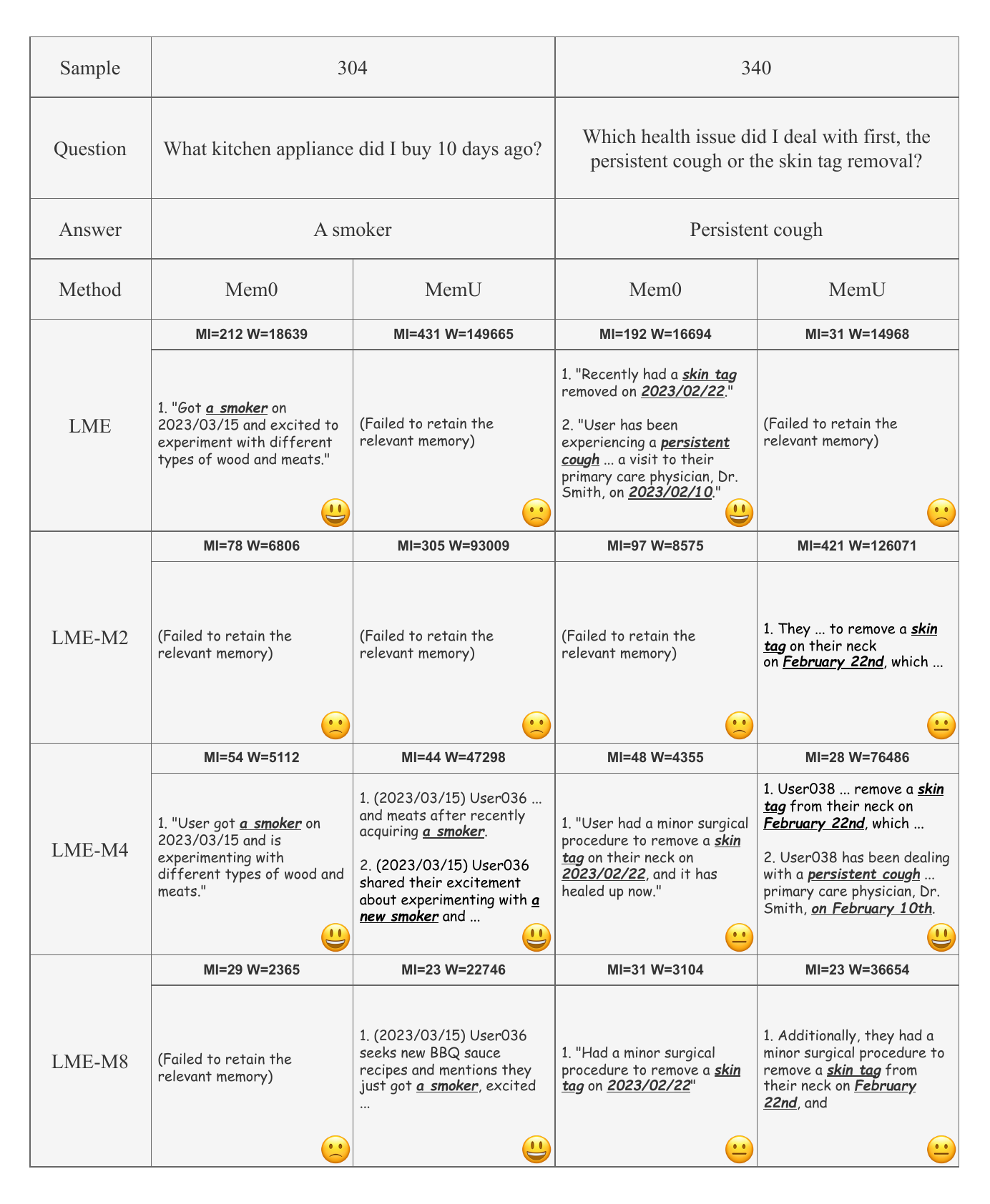}
    \caption{\textbf{Case study of uncontrolled memory retention.}
Rewrite-based memory construction does not preserve evidence monotonically across session-granularity settings. The same evidence may be covered, partially retained, or missing under different concatenation factors. The emoji markers denote \textit{Covered}, \textit{Partial}, and \textit{Missing}, respectively. MI denotes the number of memory items, and W denotes the total number of words in the constructed memory.}
\label{fig:app_uncontrolled_retention}
\end{figure}

Figure~\ref{fig:app_uncontrolled_retention} illustrates \textbf{Phenomenon 2}. The examples show that memory retention is unstable across concatenation settings. In Sample 304, the evidence that the user bought a smoker is retained in some harder settings but missing in easier ones. In Sample 340, evidence about the skin tag removal and persistent cough also appears inconsistently across settings and methods. This demonstrates that rewrite-based memory construction does not provide a controllable guarantee on evidence preservation. The three emoji markers indicate the evidence status: \textit{Covered}, \textit{Partial}, and \textit{Missing}. Even when the same dialogue content and QA target are used, changing only the session granularity can lead the memory writer to preserve, partially preserve, or completely omit the same evidence. 

\subsection{BERT Segmenter Training Details}
\label{app:segmenter_training}

\subsubsection{BERT Segmenter Training Settings}

\paragraph{Task formulation.}
We formulate incremental topic segmentation as a binary classification problem. Given the current segment buffer $s'$ and an incoming turn $t'$, the segmenter predicts whether to append $t'$ to the current segment (\texttt{CONTINUE}) or start a new segment (\texttt{SEGMENT}). Each training example therefore corresponds to a local boundary decision, matching the sequential decision process used during inference. Segmentation never splits a sentence across segments.

\paragraph{Training data construction.}
We construct segmentation supervision from the dialogue histories of abstention instances in LongMemEval. Although these instances have unanswerable questions, their histories contain multi-session user--assistant interactions with natural session boundaries and intra-session topic transitions. We use only the dialogue histories and their boundary annotations; the associated questions and answers are not used as training targets. All non-abstention evaluation instances are excluded from segmenter training.

Following Section~3.1, we combine two sources of boundary supervision. Session boundaries provide coarse-grained hard labels, while GPT-4.1-mini annotates finer-grained topic boundaries within each session. These LLM-derived annotations provide complementary weak supervision for detecting topic shifts that do not coincide with session boundaries. Both sources are converted into binary \texttt{CONTINUE}/\texttt{SEGMENT} targets.

Importantly, the number of source benchmark instances is different from the number of supervised training examples. Each source instance contains multiple sessions, and each annotated session yields a sequence of boundary-decision examples. For a continuation example, the input consists of the preceding segment context and the next turn within the same segment; for a boundary example, it consists of the preceding segment and the first turn of the next segment. Although the supervision is derived from 30 source
instances, this construction yields
\textbf{15,260 supervised boundary-decision examples}, containing
approximately 13M WordPiece tokens across the constructed inputs.
The segmenter therefore learns from thousands of local boundary
decisions despite the small number of source instances.

We partition the source histories at the dialogue level, assigning 90\% to training and 10\% to validation, so that examples derived from the same source history remain in the same split. The validation set is used for checkpoint selection and comparison of candidate encoder backbones. The original class distribution is approximately
$\texttt{CONTINUE}:\texttt{SEGMENT}=4:1$,
reflecting the greater frequency of topic continuation in the annotated conversations.

\paragraph{Encoder backbones and training protocol.}
We compare the base-size variants of BERT~\citep{devlin2019bert}, RoBERTa~\citep{liu2019roberta}, and ModernBERT~\citep{warner2025smarter} under the same binary classification formulation. For each backbone, we attach a linear classification head to the final representation of the classification token and fine-tune the entire model.

Let $y_i \in \{0,1\}$ denote the target label for example $i$, where $1$ corresponds to \texttt{SEGMENT}, and let $p_i$ denote the predicted boundary probability. We optimize the standard binary cross-entropy loss:
\begin{equation}
    \mathcal{L}_{\mathrm{CE}}
    =
    -\frac{1}{N}
    \sum_{i=1}^{N}
    \left[
        y_i \log p_i
        +
        (1-y_i)\log(1-p_i)
    \right].
    \label{eq:segmenter_ce}
\end{equation}

All backbones are trained for 5 epochs using AdamW, with a batch size of 8, a learning rate of $2 \times 10^{-5}$, and FP16 mixed precision on a single NVIDIA RTX 3090 GPU.

\paragraph{Evaluation protocol and metrics.}
We assess segmentation quality on a subset of 50 LongMemEval instances disjoint from the segmenter training data. Reference boundaries are constructed using the same annotation protocol as for training, combining session boundaries with LLM-annotated intra-session topic shifts. This subset also serves to calibrate the boundary decision threshold. Calibration uses only boundary annotations and segmentation metrics, without QA labels, gold answers, response predictions, or downstream QA accuracy.

Boundary detection accuracy alone does not fully characterize the usefulness of segmentation for memory retrieval. Excessive segmentation can fragment coherent evidence, whereas insufficient segmentation can mix unrelated topics. We therefore evaluate both boundary accuracy and segment granularity using five complementary metrics.

Let $\mathcal{D}$ denote the evaluated dialogues. For dialogue $d$, let $T_d$ be its number of turns, $P_d$ and $G_d$ its predicted and reference boundary sets, and $\mathcal{S}^{\mathrm{pred}}_d$ and $\mathcal{S}^{\mathrm{gold}}_d$ the corresponding segment partitions.

\begin{itemize}
    \item \textbf{Exact-match F1} measures agreement between predicted and reference boundaries. We compute precision and recall as
    \begin{equation}
        \mathrm{Prec}_d
        =
        \frac{|P_d \cap G_d|}{|P_d|},
        \qquad
        \mathrm{Rec}_d
        =
        \frac{|P_d \cap G_d|}{|G_d|},
    \end{equation}
    and macro-average the resulting dialogue-level F1 scores.

    \item \textbf{Window-tolerant F1 (W-F1)} allows a boundary offset of one turn. Its precision is the fraction of predicted boundaries within one turn of a reference boundary, and its recall is the fraction of reference boundaries within one turn of a predicted boundary. We report their harmonic mean.

    \item \textbf{Boundary Over-segmentation Ratio (BOR)} measures the number of predicted boundaries relative to the reference:
    \begin{equation}
        \mathrm{BOR}
        =
        \frac{\sum_{d \in \mathcal{D}} |P_d|}
             {\sum_{d \in \mathcal{D}} |G_d|}.
    \end{equation}
    Values above or below $1$ indicate over-segmentation or under-segmentation, respectively. A value close to $1$ indicates agreement in boundary count, but does not by itself establish boundary accuracy.

    \item \textbf{Purity} measures how well each predicted segment is contained within a reference segment:
    \begin{equation}
        \mathrm{Purity}_d
        =
        \frac{1}{T_d}
        \sum_{p \in \mathcal{S}^{\mathrm{pred}}_d}
        \max_{g \in \mathcal{S}^{\mathrm{gold}}_d}
        |p \cap g|.
    \end{equation}

    \item \textbf{Coverage} measures how well each reference segment is preserved within a predicted segment:
    \begin{equation}
        \mathrm{Coverage}_d
        =
        \frac{1}{T_d}
        \sum_{g \in \mathcal{S}^{\mathrm{gold}}_d}
        \max_{p \in \mathcal{S}^{\mathrm{pred}}_d}
        |g \cap p|.
    \end{equation}
\end{itemize}

\paragraph{Backbone and threshold selection.}
Table~\ref{tab:segmentation_results_final} compares the candidate backbones at representative decision thresholds. We select \textbf{BERT-base with $\tau=0.20$} as the default segmenter, considering boundary accuracy together with segment coherence and completeness. Within the BERT family, lowering the threshold to $0.15$ improves exact-match F1 but also increases over-segmentation and reduces Coverage. Raising it to $0.25$ improves boundary-count calibration and Coverage at the expense of F1 and Purity. The selected threshold provides a practical balance between these competing properties.

The selected model and threshold are fixed for all downstream experiments. In particular, we directly apply the same segmenter to PersonaMem without dataset-specific fine-tuning or threshold adjustment.

\begin{table*}[!tb]
\centering
\small
\renewcommand{\arraystretch}{1.1}
\setlength{\tabcolsep}{7pt}
\caption{
Performance comparison of varying architectures and thresholds on a randomly sampled subset of 50 evaluation instances from the LongMemEval dataset.
The arrows ($\uparrow$, $\rightarrow 1$) indicate the optimal direction for each metric.
Best and second-best results \textbf{within each model family} are highlighted in \textbf{bold} and \underline{underlined}, respectively.
The \colorbox{bestrow}{\strut highlighted row} indicates our final selected segmenter.
}
\label{tab:segmentation_results_final}
\begin{tabular}{lc ccccc}
\toprule
\multirow{2}{*}{\textbf{Base Model}}
& \multirow{2}{*}{\textbf{Threshold} ($\tau$)}
& \multicolumn{5}{c}{\textbf{LongMemEval Dataset (50 Samples)}} \\
\cmidrule(lr){3-7}
& & $F\_1$$\uparrow$ & W-$F\_1$$\uparrow$ & BOR$\rightarrow1$ & Pur.$\uparrow$ & Cov.$\uparrow$ \\
\midrule
\multicolumn{7}{c}{\textit{BERT Family}} \\
\midrule
BERT & 0.15 & \textbf{0.497} & \textbf{0.600} & 1.266 & \textbf{0.777} & 0.682 \\
\rowcolor{bestrow}
\textbf{BERT} & \textbf{0.20} & \underline{0.483} & \underline{0.594} & \underline{1.169} & \underline{0.763} & \underline{0.708} \\
BERT & 0.25 & 0.475 & 0.592 & \textbf{1.100} & 0.754 & \textbf{0.727} \\
\midrule
\multicolumn{7}{c}{\textit{RoBERTa Family}} \\
\midrule
RoBERTa & 0.15 & 0.442 & \textbf{0.658} & 1.754 & \textbf{0.852} & 0.550 \\
RoBERTa & 0.20 & \underline{0.463} & \underline{0.653} & \underline{1.630} & \underline{0.836} & \underline{0.585} \\
RoBERTa & 0.25 & \textbf{0.479} & 0.645 & \textbf{1.502} & 0.819 & \textbf{0.619} \\
\midrule
\multicolumn{7}{c}{\textit{ModernBERT Family}} \\
\midrule
ModernBERT & 0.30 & \textbf{0.482} & \textbf{0.600} & 1.093 & \textbf{0.774} & 0.697 \\
ModernBERT & 0.35 & \underline{0.478} & \underline{0.597} & \underline{1.056} & 0.768 & \underline{0.709} \\
ModernBERT & 0.40 & 0.470 & 0.594 & \textbf{1.023} & \underline{0.762} & \textbf{0.719} \\
\bottomrule
\end{tabular}
\end{table*}

\subsubsection{Controlled Training Settings}

We further analyze the effects of training data size, class distribution, and supervision mixture. Across all experiments, training hyperparameters are fixed to the default values described above, including the learning rate, batch size, number of epochs, and input-context configuration.

To obtain additional examples for the different data configurations, we supplement the candidate training pool with examples constructed from non-abstention LongMemEval dialogue histories. These histories undergo the same annotation and example-construction pipeline as the original data: session boundaries provide hard supervision, GPT-4.1-mini annotates finer-grained intra-session boundaries, and the annotated histories are converted into binary boundary-decision examples.

The training-data-size analysis varies the total number of training examples. For the class-distribution and supervision-mixture analyses, we construct training sets with the specified ratios from the candidate pool while keeping the total number of training examples identical across settings within each analysis.
\paragraph{Effect of training data size.}
To examine whether the constructed supervision is sufficient, we compare training sizes ranging from 50\% to 120\% of the original setting. For the 110\% and 120\% settings, we expand the training data with additional dialogue histories from LongMemEval. These histories undergo the same annotation and example-construction pipeline as the original data: session boundaries provide coarse-grained supervision, GPT-4.1-mini annotates intra-session topic boundaries, and the annotated histories are converted into binary boundary-decision examples.

As shown in Table~\ref{tab:segmenter_data_size}, reducing the training size to 50\% lowers F1 from 0.483 to 0.399. In contrast, F1 remains within a relatively narrow range from 80\% to 120\% of the original size. Increasing the training size beyond the original setting yields little additional benefit: the 110\% setting improves F1 by only 0.003 while increasing BOR, and the 120\% setting has both lower F1 and a higher BOR than the original configuration. These results suggest diminishing returns from additional annotated dialogues within the tested range and support the adequacy of the original training-data setting. Although some smaller settings achieve a BOR closer to $1$, they also have lower F1, highlighting the need to consider boundary accuracy and segmentation granularity jointly.

\begin{table}[t]
    \centering
    \small
    \caption{
        Effect of training data size on segmentation quality.
        Training sizes are expressed relative to the original setting.
        The original configuration is shown in bold.
    }
    \label{tab:segmenter_data_size}
    \begin{tabular}{lcc}
        \toprule
        Relative training size & F1$\uparrow$ & BOR$\rightarrow 1$ \\
        \midrule
        50\%  & 0.399 & 0.914 \\
        80\%  & 0.471 & 1.114 \\
        90\%  & 0.454 & 1.054 \\
        \textbf{100\% (original)} & \textbf{0.483} & \textbf{1.169} \\
        110\% & 0.486 & 1.247 \\
        120\% & 0.478 & 1.205 \\
        \bottomrule
    \end{tabular}
\end{table}

\paragraph{Effect of class distribution.}
We further examine the influence of the
$\texttt{CONTINUE}:\texttt{SEGMENT}$ ratio while keeping the total number of training examples identical across all settings. The original training data have an approximately $4:1$ ratio, reflecting the greater frequency of topic continuation in the annotated interaction streams. To construct equal-sized training sets with different class ratios, we supplement the candidate training pool with examples derived from additional non-abstention LongMemEval dialogue histories, using the same boundary-annotation and example-construction pipeline. All training hyperparameters remain fixed across these settings.

Table~\ref{tab:segmenter_class_ratio} shows that all tested alternatives to the original ratio produce more boundaries relative to the reference. Equalizing the classes produces a BOR of 2.228, indicating more than twice as many predicted boundaries as reference boundaries. Less aggressive changes also increase over-segmentation. In particular, the $3:1$ setting achieves nearly the same F1 as the original configuration but produces substantially more boundaries.

Among the tested class ratios, the original $4:1$ distribution achieves the BOR closest to $1$ while maintaining competitive F1. Under the same training-data size and fixed training hyperparameters, these results support retaining the observed class distribution to balance boundary detection accuracy with the preservation of coherent evidence segments.

\begin{table}[t]
    \centering
    \small
    \caption{
        Effect of the training class ratio on segmentation quality.
        C and S denote \texttt{CONTINUE} and \texttt{SEGMENT},
        respectively. The original configuration is shown in bold.
    }
    \label{tab:segmenter_class_ratio}
    \begin{tabular}{lcc}
        \toprule
        Class ratio (C:S) & F1$\uparrow$ & BOR$\rightarrow 1$ \\
        \midrule
        $1:1$ & 0.421 & 2.228 \\
        $2:1$ & 0.477 & 1.276 \\
        $3:1$ & 0.484 & 1.511 \\
        \textbf{$4:1$ (original)} & \textbf{0.483} & \textbf{1.169} \\
        \bottomrule
    \end{tabular}
\end{table}

\paragraph{Effect of supervision mixture.}
We assess sensitivity to the composition of hard and soft boundary supervision. Hard labels correspond to explicit session boundaries, while LLM-annotated soft labels capture finer-grained intra-session topic shifts. In the original training set, the hard-to-soft label ratio is $0.902:1$, approximately $1:1$. Here, Hard:Soft refers to the relative contributions of the two supervision sources, rather than the \texttt{CONTINUE}:\texttt{SEGMENT} class ratio. All training hyperparameters remain fixed across the compared settings.

As shown in Table~\ref{tab:segmenter_supervision_mix}, hard-only supervision produces conservative boundary predictions, resulting in low F1 and substantial under-segmentation. Incorporating soft labels improves boundary detection, but soft-only supervision leads to pronounced over-segmentation. The original mixture achieves both the highest F1 and the BOR closest to $1$ among the tested configurations. These results support the complementary roles of the two supervision sources: explicit session boundaries provide coarse structural anchors, while LLM annotations supply finer-grained topic-transition signals. Their combination provides a favorable balance between boundary detection accuracy and segmentation granularity.

\begin{table}[t]
    \centering
    \small
    \caption{
        Effect of the hard/soft supervision mixture on segmentation
        quality under fixed training hyperparameters.
        The original hard-to-soft ratio is $0.902:1$.
        The original configuration is shown in bold.
    }
    \label{tab:segmenter_supervision_mix}
    \begin{tabular}{lcc}
        \toprule
        Supervision setting & F1$\uparrow$ & BOR$\rightarrow 1$ \\
        \midrule
        Hard-only & 0.226 & 0.505 \\
        Hard:Soft $=2:1$ & 0.321 & 0.674 \\
        Hard:Soft $=1:2$ & 0.459 & 0.816 \\
        Soft-only & 0.472 & 1.907 \\
        \textbf{Original mix ($0.902:1$)}
        & \textbf{0.483} & \textbf{1.169} \\
        \bottomrule
    \end{tabular}
\end{table}

Together, these analyses support the adequacy of the original training-data setting and the complementary value of the two supervision sources. Performance remains relatively stable under moderate changes in training data size, whereas changes in supervision mixture or class distribution can substantially affect segmentation granularity. With training hyperparameters held fixed, the original configuration provides a practical balance between boundary detection accuracy and boundary-count calibration.

\subsection{LLM-As-A-Judge Details}
\label{app:judge}

While LLM-as-a-Judge has become a common practice for long-term dialogue memory evaluation, we observe that a simple binary grading scheme can be overly permissive in this setting. 
When the judge only decides whether an answer is ``correct'' or ``incorrect'', vague, under-specified, or weakly supported predictions may be accepted as correct as long as they partially overlap with the gold answer. 
This is especially problematic for memory QA, where the key challenge is to recover the specific user-related fact, temporal relation, or preference evolution required by the question.

To obtain a more reliable evaluation, we use GPT-4.1-mini as the judge model and adopt a stricter four-way protocol: \textit{correct}, \textit{partially\_correct}, \textit{incorrect}, and \textit{insufficient}. 
The \textit{correct} label is used only when the prediction clearly matches the core fact in the gold answer, allowing for paraphrases, equivalent date formats, and non-conflicting extra details. 
The \textit{partially\_correct} label is used when the prediction contains relevant but incomplete information, while \textit{incorrect} is used for wrong, contradictory, unrelated, or conflicting predictions. 
The \textit{insufficient} label is used for answers that are too vague, evasive, or unverifiable. 
Only predictions labeled as \textit{correct} are counted as correct, reducing false positives from incomplete or ambiguous answers.

\begin{promptbox}{Judge Prompt}
You are an expert grader for memory QA.
Your job is to compare a model prediction with a gold answer and decide whether they match on the core fact.

Grading policy (important):
1) Be reasonably generous, not strict string-match.
2) If prediction clearly expresses the same core fact/topic as the gold answer, mark it correct even if wording differs.
3) Prediction may be longer than gold; extra non-conflicting detail is allowed.
4) For time answers, treat equivalent formats/references as correct (e.g., "May 7" vs "7 May"; relative wording that maps to same date/period).
5) Mark incorrect only when the core fact is wrong, missing, contradictory, or mostly unrelated.
6) Use partially_correct only when prediction contains a relevant but incomplete core answer.
7) Penalize contradictions to gold; if prediction includes both correct and incorrect conflicting claims, mark incorrect.

Return JSON only with keys:
- is_correct: boolean
- verdict: one of ["correct", "partially_correct", "incorrect", "insufficient"]
- reason: short explanation

Calibration:
- For verdict "correct", set is_correct=true.
- For verdict "partially_correct", set is_correct=false.
- For verdict "incorrect" or "insufficient", set is_correct=false.
\end{promptbox}

\subsection{Empirical Results on Qwen-family Models}
\label{app:qwen}

To examine whether the observed improvements depend on a specific model family, we further conduct a cross-backbone evaluation using Qwen-family models. In the main experiments, we use GPT-4.1-mini as the general memory construction model and GPT-5-mini as the response model. In this appendix setting, we replace them with Qwen3.5-Flash and Qwen3.5-Plus, respectively. This setting changes both the memory-writing model and the response generator, and therefore should be interpreted as a cross-backbone robustness test rather than a component-wise ablation.

The two Qwen models play different roles in our pipeline. Qwen3.5-Flash is used as the general memory construction model, corresponding to GPT-4.1-mini in the main setting. It is designed as a faster and more cost-efficient model, which makes it suitable for large-scale memory construction but may be less stable for fine-grained memory extraction, summarization, and update operations. Qwen3.5-Plus is used as the response model, corresponding to GPT-5-mini in the main setting. Compared with Qwen3.5-Flash, it is a stronger model in the Qwen3.5 family and supports thinking-enabled response generation. This makes it particularly suitable for reasoning-oriented personalization tasks, but also increases query-time latency. Therefore, this appendix setting provides a useful stress test for both construction-time robustness and response-time reasoning ability.

Table~\ref{tab:longmemeval_qwen} and Table~\ref{tab:personamem_qwen} report the cross-backbone results with Qwen-family models. 
Overall, \texttt{Threader} remains the strongest overall method on both datasets, demonstrating that the advantage of structure-aware memory access is not tied to a specific model family. 
On LongMemEval, \texttt{Threader} achieves an overall accuracy of 87.00, outperforming the strongest baseline, EverMemOS, by 8.21\% relative improvement. 
On PersonaMem, \texttt{Threader} also obtains the best overall score of 82.00, outperforming the strongest baseline, A-Mem, by 2.98\% relative improvement. 

\begin{table*}[!tb]
\caption{\textbf{Main results on LongMemEval with Qwen-family models.}
\texttt{Threader} retrieves the top-12 segments for LongMemEval.
SS-User, SS-Assi., SS-Pref., Multi-S., Know.-U., and Temp.-R. denote
Single-Session-User, Single-Session-Assistant, Single-Session-Preference,
Multi-Session, Knowledge-Update, and Temporal-Reasoning, respectively.
Results are grouped by backbone. Within each backbone block, the best
result in each column is shown in \textbf{bold}, the second-best
distinct result is \underline{underlined}.}
\label{tab:longmemeval_qwen}

\centering
\footnotesize
\renewcommand{\arraystretch}{1.0}
\setlength{\tabcolsep}{1.8pt}

\begin{tabular}{@{}lccccccc@{}}
\toprule
\textbf{Method}
& \textbf{SS-User}
& \textbf{SS-Assi.}
& \textbf{SS-Pref.}
& \textbf{Multi-S.}
& \textbf{Know.-U.}
& \textbf{Temp-R.}
& \textbf{Overall} \\
\midrule

\multicolumn{8}{c}{\textit{Qwen3.5-plus}} \\

Self-RAG
& 90.00 & 94.64 & 60.00 & 69.17 & 79.49 & 80.45 & 79.00 \\

HippoRAG2
& 77.14 & \underline{96.43} & \underline{80.00}
& 72.93 & \underline{88.46} & 71.43 & 78.60 \\

Mem0
& 72.86 & 21.43 & 26.67 & 40.60 & 48.72 & 56.39 & 47.60 \\

MemU
& \textbf{98.57} & 82.14 & 40.00 & \textbf{82.71}
& 87.18 & 45.86 & 73.20 \\

A-Mem
& 92.86 & \textbf{98.21} & 26.67 & 54.89
& 78.21 & 64.66 & 69.60 \\

MemOS
& 81.43 & 89.29 & \textbf{90.00}
& 60.15 & 83.33 & 47.37 & 68.40 \\

EverMemOS
& 94.29 & 71.43 & 66.67 & \underline{75.94}
& 83.33 & \underline{82.71} & \underline{80.40} \\

\rowcolor{gray!25}
\textbf{\texttt{Threader}}
& \underline{97.14}\,\mbox{\tiny($\downarrow$1.45\%)}
& \textbf{98.21}\,\mbox{\tiny($\uparrow$0.00\%)}
& 70.00\,\mbox{\tiny($\downarrow$22.22\%)}
& \underline{75.94}\,\mbox{\tiny($\downarrow$8.18\%)}
& \textbf{93.59}\,\mbox{\tiny($\uparrow$5.80\%)}
& \textbf{87.97}\,\mbox{\tiny($\uparrow$6.36\%)}
& \textbf{87.00}\,\mbox{\tiny($\uparrow$8.21\%)} \\

\bottomrule
\end{tabular}

\vspace{-1.0em}
\end{table*}

\begin{table*}[!tb]
\caption{\textbf{Main results on PersonaMem with Qwen-family models.}
\texttt{Threader} retrieves the top-8 segments for PersonaMem.
Generalize-New-Scn., Preference-Align, User-Shared, User-Mentioned,
Update-Reason, New-Ideas, and Preference-Evolution denote
generalizing-to-new-scenarios, provide-preference-aligned-recommendations,
recalling-user-shared-facts, recalling-facts-mentioned-by-the-user,
recalling-reasons-behind-previous-updates, suggesting-new-ideas, and
tracking-full-preference-evolution, respectively.
Results are grouped by backbone. Within each backbone block, the best
result in each column is shown in \textbf{bold}, and the second-best
distinct result is \underline{underlined}.}
\label{tab:personamem_qwen}

\centering
\footnotesize
\renewcommand{\arraystretch}{1.0}
\setlength{\tabcolsep}{1pt}

\begin{tabular}{@{}lcccccccc@{}}
\toprule
\multirow{2}{*}{\textbf{Method}}
& \textbf{Generalize}
& \textbf{Preference}
& \textbf{User-}
& \textbf{User-}
& \textbf{Update-}
& \textbf{New-}
& \textbf{Preference}
& \multirow{2}{*}{\textbf{Overall}} \\
& \textbf{-New-Scn.}
& \textbf{-Align}
& \textbf{Shared}
& \textbf{Mentioned}
& \textbf{Reason}
& \textbf{Ideas}
& \textbf{Evolution}
& \\
\midrule

\multicolumn{9}{c}{\textit{Qwen3.5-plus}} \\

Self-RAG
& 29.82 & 63.64 & 10.85 & 52.94
& 77.78 & \textbf{59.14} & 72.66 & 52.29 \\

HippoRAG2
& 78.95 & 80.00 & 79.84 & 58.82
& 83.84 & 46.24 & 74.10 & 73.17 \\

Mem0
& 64.91 & 54.55 & 37.21 & 58.82
& 83.84 & 39.78 & \underline{86.33} & 61.97 \\

MemU
& 61.40 & 72.73 & \underline{81.39} & 64.71
& 85.86 & 38.71 & 76.26 & 70.97 \\

A-Mem
& \underline{91.23} & \textbf{90.91} & 73.64 & \textbf{82.35}
& \textbf{95.96} & 50.54 & 83.45 & \underline{79.63} \\

MemOS
& 78.95 & 81.82 & 75.19 & 70.59
& 85.86 & 37.63 & 76.98 & 72.33 \\

EverMemOS
& \underline{91.23} & \underline{83.64} & \textbf{82.17} & 64.71
& \underline{91.92} & 35.48 & 84.89 & 77.59 \\

\rowcolor{gray!25}
\textbf{\texttt{Threader}}
& \textbf{94.74}\,\mbox{\vtiny($\uparrow$3.85\%)}
& \textbf{90.91}\,\mbox{\vtiny($\rightarrow$0.00\%)}
& \textbf{82.17}\,\mbox{\vtiny($\rightarrow$0.00\%)}
& \underline{76.47}\,\mbox{\vtiny($\downarrow$7.14\%)}
& 89.90\,\mbox{\vtiny($\downarrow$6.32\%)}
& \underline{52.69}\,\mbox{\vtiny($\downarrow$10.91\%)}
& \textbf{87.77}\,\mbox{\vtiny($\uparrow$1.67\%)}
& \textbf{82.00}\,\mbox{\vtiny($\uparrow$2.98\%)} \\

\bottomrule
\end{tabular}

\vspace{-1.5em}
\end{table*}
On LongMemEval, the gains remain most pronounced on evidence-intensive question types. 
\texttt{Threader} achieves the best results on \textit{Knowledge-Update} and \textit{Temporal-Reasoning}, with scores of 93.59 and 87.97, respectively. 
These categories require locating fine-grained historical evidence and preserving update or temporal relations, where lossy memory construction can be sensitive to the quality of the memory-writing model. 
This sensitivity is reflected in the performance of rewrite-heavy memory systems: Mem0 drops from 57.20 in the GPT-based setting to 47.60 in the Qwen-based setting, and EverMemOS drops from 84.80 to 80.40. 
Since these methods rely on the construction model to extract, summarize, merge, and update memory entries, replacing GPT-4.1-mini with Qwen3.5-Flash can affect the quality of the generated memory bank. 
In contrast, \texttt{Threader} avoids construction-time LLM rewriting and preserves raw dialogue evidence through topic-coherent segmentation, making its memory construction process less dependent on the summarization ability of a particular backbone. 
Although \texttt{Threader} also slightly decreases from 90.60 to 87.00, it preserves its overall advantage and remains strongest on the most evidence-intensive categories.

PersonaMem shows a different pattern.
Under the Qwen backbone, multiple methods achieve substantially higher overall accuracy than in the GPT-based setting.
For example, HippoRAG2 improves from 59.42 to 73.17, EverMemOS from 65.20 to 77.59, A-Mem from 60.78 to 79.63, and \texttt{Threader} from 69.95 to 82.00.
These gains suggest that backbone choice plays an important role in PersonaMem performance.
Many PersonaMem questions require reasoning about user preferences, tracking their evolution, and generalizing them to new scenarios.
The improvements may therefore partly reflect Qwen3.5-Plus's ability to interpret and integrate retrieved evidence for these reasoning tasks.

Under the Qwen backbone, \texttt{Threader} achieves the best results on \textit{generalizing-to-new-scenarios}, \textit{provide-preference-aligned-recommendations}, \textit{recall-user-shared-facts}, and \textit{tracking-full-preference-evolution}. 
This further shows that topic-coherent raw evidence retrieval can be effectively exploited by a different model family, especially when the response model has stronger reasoning ability. 
However, \texttt{Threader} is less advantageous on \textit{recalling-the-reasons-behind-previous-updates}, \textit{suggest-new-ideas}, and questions, where explicit abstraction of preference changes or higher-level user profiles can be beneficial. 
This observation is consistent with the trade-off observed in the main experiments: \texttt{Threader} prioritizes faithful evidence preservation and retrieval, while proactive preference abstraction remains complementary.
 
Overall, the Qwen-based results indicate that \texttt{Threader} is robust across model families and particularly effective when memory access depends on preserving distributed evidence. 
At the same time, they highlight an accuracy--efficiency trade-off: thinking-enabled response generation improves reasoning-heavy personalization performance, but substantially increases online serving latency.

\subsection{Additional Ablation Study}

\subsubsection{Ablation Study of Multi-view representation in the full pipeline.}
The dense-only ablation isolates the contribution of multi-view segment representations. We further examine whether this contribution remains beneficial when BM25, message-level dense scoring, and cross-encoder reranking are enabled. Multi-view representation primarily affects Stage~I candidate retrieval, while Stage~II refines the retrieved candidates. Its effect on final accuracy therefore depends on the candidate pool size $|\mathcal{C}|$. In the main setting, the max candidate pool size is 80 ($2\times K_1$), which may allow relevant segments to enter Stage~II both with and without multi-view representation, despite differences in their initial ranks. Subsequent refinement can then recover these segments in either setting, making final accuracy less sensitive to the quality of Stage~I ranking.

To make the contribution of candidate selection more apparent, we reduce $K_1$ from 40 to 10 and compare the complete pipeline with a variant that removes multi-view representation at Top-5. Both variants retain BM25 retrieval, message-level dense scoring, and cross-encoder reranking. As shown in Table~\ref{tab:full_pipeline_multiview}, multi-view representation improves accuracy from 78.4\% to 81.2\%, a gain of 2.8 percentage points. This result demonstrates its complementary value within the full retrieval pipeline under a more restrictive candidate budget: improving candidate selection remains important because downstream refinement can only operate on evidence admitted to Stage~II.

\begin{table}[t]
    \centering
    \small
    \caption{
        \textbf{Full-pipeline ablation of multi-view representation on LongMemEval}, with
        $K_1$ reduced from 40 to 10.
        Both variants retain BM25 retrieval, message-level dense
        scoring, and cross-encoder reranking.
    }
    \label{tab:full_pipeline_multiview}
    \begin{tabular}{lc}
        \toprule
        Method & Accuracy (\%) \\
        \midrule
        \textsc{Threader} w/o multi-view representation & 78.4 \\
        \textsc{Threader} (full pipeline) & \textbf{81.2} \\
        \bottomrule
    \end{tabular}
\end{table}

\begin{table*}[t]
\centering
\small
\setlength{\tabcolsep}{8pt}

\caption{\textbf{Ablation study on the key design axes of \texttt{Threader} on Personamem.}
We evaluate the impact of segmentation, multi-view representation,
retrieval signals, and retrieval granularity on answer accuracy.
Emb.-User, Emb.-Asst., and Emb.-Full correspond to using only user,
assistant, or full interaction views, respectively.}
\label{tab:ablation_personamem}

\begin{tabular}{@{}c@{\hspace{5mm}}|@{\hspace{5mm}}c@{}}

\begingroup
\renewcommand{\arraystretch}{1.191667}
\begin{tabular}[t]{lc}
\toprule
\textbf{Variant} & \textbf{ACC} \\
\midrule

\multicolumn{2}{l}{
    \textcolor{teal}{\textit{\textbf{Segmentation ablation}}}
} \\

\rowcolor{gray!25}
\texttt{Threader} & 69.95 \\

Chunk-300. & 65.70 \\
Chunk-500  & 67.40 \\
Chunk-700  & 63.67 \\
Chunk-1K   & 59.76 \\

\cmidrule(lr){1-2}

\multicolumn{2}{l}{
    \textcolor{teal}{\textit{\textbf{Multi-view representation ablation}}}
} \\

\rowcolor{gray!25}
Emb. w/o Msg. & 58.40 \\

Emb.-Full  & 58.23 \\
Emb.-User  & 58.23 \\
Emb.-Asst. & 54.67 \\

\bottomrule
\end{tabular}
\endgroup

&

\begingroup
\renewcommand{\arraystretch}{1.1}
\begin{tabular}[t]{lc}
\toprule
\textbf{Variant} & \textbf{ACC} \\
\midrule

\multicolumn{2}{l}{
    \textcolor{teal}{\textit{\textbf{Retrieval signal ablation}}}
} \\

\rowcolor{gray!25}
\texttt{Threader} & 69.95 \\

w/o Emb.          & 67.40 \\
w/o BM25          & 65.87 \\
w/o Rerank.       & 66.04 \\
w/o Emb.\&BM25    & 62.82 \\
w/o Emb.\&Rerank. & 66.55 \\
w/o BM25\&Rerank. & 62.14 \\
w/o Msg.-Emb.     & 65.70 \\

\cmidrule(lr){1-2}

\multicolumn{2}{l}{
    \textcolor{teal}{\textit{\textbf{Retrieval granularity ablation}}}
} \\

\rowcolor{gray!25}
Segment Unit & 69.95 \\

Message Unit & 67.06 \\

\bottomrule
\end{tabular}
\endgroup

\end{tabular}
\end{table*}

\subsubsection{Ablation Study on PersonaMem}
Table~\ref{tab:ablation_personamem} reports ablation results on PersonaMem along the same four design axes as the main experiments: segmentation, multi-view representation, retrieval signals, and retrieval granularity.

\paragraph{Segmentation ablation and cross-dataset transfer.}
Replacing the learned topic-coherent segmentation with fixed-size chunking consistently reduces answer accuracy. \textsc{Threader} achieves 69.95\%, outperforming the strongest fixed-size alternative, Chunk-500, by 2.55 percentage points. Notably, the segmenter is trained exclusively on LongMemEval dialogue histories and \emph{directly transferred to PersonaMem}, with the same model parameters and decision threshold, without dataset-specific fine-tuning or threshold adjustment. Despite the substantial differences in session length and dialogue structure between the two benchmarks, the transferred segmenter remains effective for downstream memory retrieval. These results support the cross-dataset generalization of the segmentation approach and show that its benefits extend beyond the benchmark used for training.

\paragraph{Representation and retrieval ablations.}
The remaining ablations broadly align with the findings on LongMemEval. Combining segment views performs competitively with or better than individual views, although the gains over user-only and full-interaction representations are modest. Removing retrieval signals or message-level refinement degrades performance, supporting the value of combining complementary relevance signals and localized evidence matching. Retrieving individual messages instead of complete segments also reduces accuracy, highlighting the importance of preserving contextual coherence. Together, these results support the effectiveness of \textsc{Threader}'s representation and retrieval design across the two benchmarks.

\subsection{Retrieval Hyperparameter Sensitivity}
\label{app:hyper}
We examine how changes in the retrieval weights affect evidence recall in \texttt{Threader}. We calibrate the segmentation threshold and retrieval weights on the same 50-instance LongMemEval subset used for segmenter evaluation in Appendix~\ref{app:segmenter_training}, using segmentation quality and evidence recall, respectively. We then freeze the trained segmenter and these calibrated settings and apply them unchanged to PersonaMem, without further fine-tuning or calibration. Our main configuration uses $(w_u,w_a,w_f)=(0.5,0.25,0.25)$, $\beta=0.65$, and $\alpha=0.3$. The analyses below characterize sensitivity to variations in these weights.

The view weights $(w_u,w_a,w_f)$ combine query--segment cosine similarities for the user-only, assistant-only, and full-interaction representations in Eq.~\ref{eq:dense}. The parameter $\beta$ controls the contribution of BM25 relative to the refined message-level dense score in Eq.~\ref{eq:hybrid}. The parameter $\alpha$ balances this hybrid score with the segment-level cross-encoder reranking score in Eq.~\ref{eq:final}. Unless otherwise specified, we fix the coarse-retrieval budget to $K_1=40$, and the message-level scoring budget to $K_2=2$.

\begin{table}[!tb]
\centering
\small
\setlength{\tabcolsep}{4.5pt}
\renewcommand{\arraystretch}{1.08}
\caption{\textbf{Segment-view weight analysis.}
We evaluate different dense segment-view weights $(w_u,w_a,w_f)$ using segment-level dense retrieval only, with BM25 matching and message-level dense matching disabled. 
Full-R denotes the fraction of queries with complete evidence coverage, and Mean-R denotes average evidence recall. }
\label{tab:query_weight_analysis}
\begin{tabular}{c cc cc cc cc}
\toprule
\multirow{2}{*}{$(w_u,w_a,w_f)$}
& \multicolumn{2}{c}{\textbf{Top-5}}
& \multicolumn{2}{c}{\textbf{Top-8}}
& \multicolumn{2}{c}{\textbf{Top-10}}
& \multicolumn{2}{c}{\textbf{Top-12}} \\
\cmidrule(lr){2-3}
\cmidrule(lr){4-5}
\cmidrule(lr){6-7}
\cmidrule(lr){8-9}
& Full-R & Mean-R
& Full-R & Mean-R
& Full-R & Mean-R
& Full-R & Mean-R \\
\midrule
$(0.3,0.35,0.35)$   & 0.6596 & 0.7644 & 0.7638 & 0.8464 & 0.8064 & 0.8758 & 0.8489 & 0.9041 \\
$(0.4,0.3,0.3)$     & 0.6660 & 0.7666 & \textbf{0.7723} & \textbf{0.8541} & 0.8128 & 0.8807 & 0.8617 & 0.9129 \\
$(0.5,0.25,0.25)$   & \textbf{0.6745} & \textbf{0.7734} & \textbf{0.7723} & 0.8534 & 0.8170 & 0.8817 & \textbf{0.8681} & \textbf{0.9165} \\
$(0.6,0.2,0.2)$     & 0.6553 & 0.7625 & \textbf{0.7723} & 0.8534 & 0.8213 & 0.8832 & 0.8638 & 0.9141 \\
$(0.7,0.15,0.15)$   & 0.6511 & 0.7509 & 0.7702 & 0.8524 & \textbf{0.8277} & \textbf{0.8856} & 0.8596 & 0.9141 \\
\bottomrule
\end{tabular}
\end{table}

\paragraph{Segment-view Weight Sensitivity.}
We isolate the effect of segment-view weights using segment-level dense retrieval alone, with BM25 matching and message-level dense matching disabled. Table~\ref{tab:query_weight_analysis} shows that increasing the user-only view weight does not monotonically improve evidence recall. The relative performance of the weighting schemes also depends on the retrieval budget: $(0.7,0.15,0.15)$ performs best at Top-10, whereas the main configuration $(0.5,0.25,0.25)$ performs best at Top-5 and Top-12 and remains competitive at Top-8. Across the five evaluated configurations, Top-12 Full-R ranges from 84.89\% to 86.81\%, and Mean-R from 90.41\% to 91.65\%. These results show a budget-dependent trade-off between emphasizing user-side evidence and retaining complementary assistant-side and full-interaction signals, with relatively modest variation at Top-12.

\begin{figure}
    \centering
    \includegraphics[width=\linewidth]{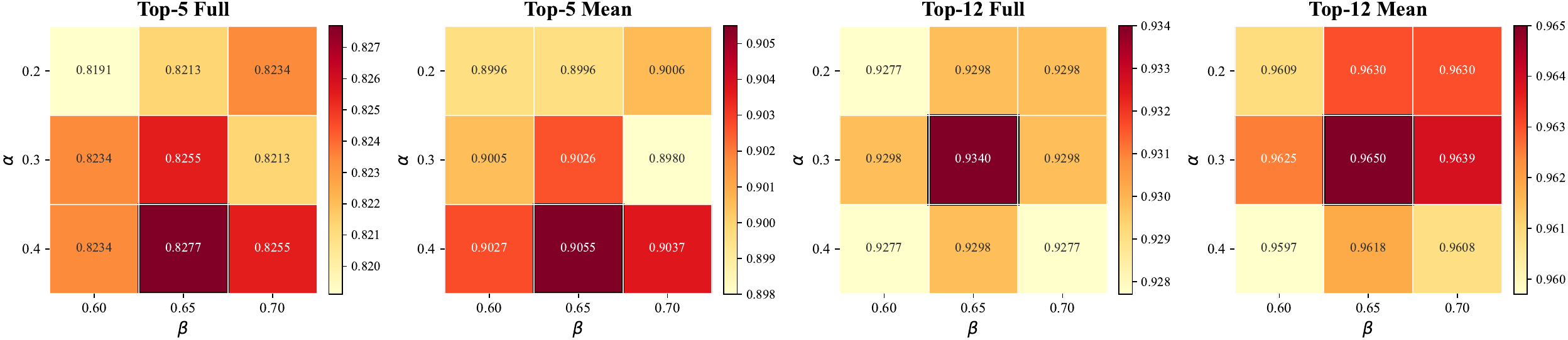}
    \caption{\textbf{Fusion weight sensitivity analysis.}
Performance heatmaps of different BM25 interpolation weights $\beta$ and final fusion weights $\alpha$ under Top-5 and Top-12 retrieval settings.}
    \label{fig:beta_alpha_heatmap}
\end{figure}

\paragraph{Fusion Weight Sensitivity.}
Holding the segment-view weights at $(0.5,0.25,0.25)$, we vary $\beta\in\{0.60,0.65,0.70\}$ and $\alpha\in\{0.2,0.3,0.4\}$. Figure~\ref{fig:beta_alpha_heatmap} reports evidence recall across this local grid. At Top-12, Full-R ranges from 92.77\% to 93.40\%, and Mean-R from 95.97\% to 96.50\%, corresponding to spans of 0.63 and 0.53 percentage points. At Top-5, the corresponding spans are 0.86 and 0.75 percentage points. Thus, both recall metrics vary by less than one percentage point across the nine displayed configurations at each retrieval budget, indicating limited sensitivity to small fusion-weight changes within the examined range.

\subsection{Abstraction and Preference-Based Tasks}
\label{app:abstraction_analysis}

Preserving conversational evidence and constructing abstract memories involve different design trade-offs. \texttt{Threader} retains original interactions and defers higher-level synthesis to response generation, whereas rewrite-based methods perform part of this abstraction during memory construction. Consequently, tasks involving preference inference, preference evolution, or personalized idea generation may depend strongly on the response model's ability to synthesize retrieved evidence. We investigate this issue through evidence inspection and a controlled comparison of response models.

\paragraph{Diagnosing retrieval and generation errors.}
We investigate whether \texttt{Threader}'s limitations on preference-related tasks arise from missing evidence during retrieval or from insufficient synthesis during response generation. We manually inspect the evidence retrieved for LongMemEval single-session preference questions and observe 100\% evidence recall. Nevertheless, answer accuracy with GPT-5-mini is 96.67\%. Thus, the remaining errors on this task occur despite the availability of the required evidence, identifying response generation as the bottleneck.

\paragraph{Improving generation with the same retrieved evidence.}
To further examine this distinction, we keep each method's memory bank, retrieved evidence, and response prompt unchanged, and replace GPT-5-mini with GPT-5.6-sol using maximum reasoning effort. As shown in Table~\ref{tab:abstraction_response_models}, \texttt{Threader}'s accuracy across the three selected tasks increases from 60.69\% to 75.95\%, achieving the highest overall score among the compared methods. Its single-session preference accuracy reaches 100\%, and it also achieves the best performance on New-Ideas. Because these improvements are obtained from the same retrieved evidence, they demonstrate that the preserved memory units contain information that a more capable response model can use more effectively.

\paragraph{The importance of evidence preservation.}
These results highlight the distinction between retrieving sufficient evidence and successfully synthesizing it into an answer. The evidence audit directly identifies a generation bottleneck on single-session preference questions, while the controlled response-model comparison demonstrates the substantial role of generation across the broader set of preference-related tasks. Crucially, preserving the original evidence allows stronger downstream reasoning to improve answers without reconstructing the memory bank or retrieving different content. This supports \texttt{Threader}'s design principle: retaining fine-grained evidence provides a reliable basis for query-time abstraction. Combining this evidence-preserving foundation with lightweight summaries remains a complementary direction, as discussed in Appendix\ref{app:limitation}.

\begin{table}[t]
    \centering
    \small
    \setlength{\tabcolsep}{4pt}
    \caption{
        Accuracy (\%) on three selected abstraction-related question types.
        SS-Pref.\ is from LongMemEval; New-Ideas and Preference Evolution
        are from PersonaMem. Overall denotes accuracy pooled across the
        262 questions in these three categories.
        GPT-5.6-sol uses maximum reasoning effort.
        The best result in each metric is shown in bold.
    }
    \label{tab:abstraction_response_models}
    \begin{tabular}{llcccc}
        \toprule
        Memory method & Response model & SS-Pref. & New-Ideas
        & \shortstack{Preference\\Evolution} & Overall \\
        \midrule
        Mem0 & GPT-5-mini
        & 26.67 & 27.96 & 66.91 & 48.47 \\
        Mem0 & GPT-5.6-sol
        & 46.67 & 48.39 & 58.99 & 53.82 \\
        \midrule
        EverMemOS & GPT-5-mini
        & 96.67 & 26.88 & 64.03 & 54.58 \\
        EverMemOS & GPT-5.6-sol
        & \textbf{100.00} & 59.14 & \textbf{79.14} & 74.43 \\
        \midrule
        \textbf{\texttt{Threader}} & GPT-5-mini
        & 96.67 & 45.16 & 63.31 & 60.69 \\
        \textbf{\texttt{Threader}} & GPT-5.6-sol
        & \textbf{100.00} & \textbf{65.59} & 77.70
        & \textbf{75.95} \\
        \bottomrule
    \end{tabular}
\end{table}

\section{Response Prompt}
\label{app:response prompt}
\begin{promptbox}{Response Prompt for LongMemEval}
You are a response agent. You will receive:
1) a user question
2) a retrieved History Session (time-ordered, possibly multi-session).

Answer using ONLY the History Session. Do not invent facts.

Question: {question}

History Session:
{history_session}

Rules:
1) Use only supported evidence from History Session; combine evidence across segments/sessions when needed.
2) If facts conflict over time, use the most recent supported fact unless the question explicitly asks for history/comparison.
3) Keep names, numbers, dates, places, and units exact.
4) For count/sum/difference/time-window questions, first apply the temporal condition (before/after/within-last-X), then compute.
5) If Question contains "[Question Date: ...]", use it as the anchor for relative time expressions; otherwise use the best explicit reference time in History Session.
6) If evidence is sufficient, answer directly. If insufficient for a required value, state the missing key fact (do not speculate).
7) For preference-related questions (including suggestion/tips), prioritize user-specific established preferences and constraints from History Session; avoid generic advice that is not grounded in history.
8) Do not claim "not mentioned/no history" when relevant evidence exists in History Session.
9) Output exactly one final answer target. Do not provide multiple candidates, unnecessary alternatives, or contrastive add-ons.
10) If ordering is requested, return an ordered list (1., 2., 3., ...); otherwise keep a single concise answer sentence.
11) For preference questions, use:
Preferred: <core preference>. Avoid/Constraints: <constraints/non-preferences>.

Output:
- Output ONLY the final answer (no analysis, no tags).
\end{promptbox}

\begin{promptbox}{Response Prompt for PersonaMem}
You are solving a PersonaMem multiple-choice memory task.

Inputs:
1) Retrieved conversation history (memory evidence)
2) Current user question
3) Four answer options

Task:
Pick the ONE option (A/B/C/D) best supported by user-specific memory.

Internal Decision Rules:
1) Build a compact memory profile from history: likes/dislikes, constraints, recent updates, and reasons for changes (earlier -> later).
2) Score each option by:
   a) Memory fit: specific match to user facts/preferences/constraints.
   b) Temporal fit: respects latest state and correct change direction when evolution is asked.
   c) Transfer quality (for recommendation/generalizing): applies the user's stable patterns (social intensity, effort tolerance, pace, style, context) to the new scenario.
   d) Conflict penalty: contradicts explicit dislikes, failed experiences, or later updates.
   e) Unsupported-detail penalty: adds specific entities/claims not evidenced in history.
3) Tie-break: prefer the option with stronger personalized evidence and fewer unsupported assumptions; do not prefer generic but plausible advice.
4) For recall-style questions, exact factual consistency is primary; for recommendation/generalizing questions, preference-and-constraint transfer is primary.

Hard Constraints:
- Use only retrieved history; no outside knowledge guessing.
- This is option selection, not open-ended answering.
- Output one letter only.

Output Format (strict):
- A single uppercase letter: A, B, C, or D
- Nothing else

[Retrieved History]
{history_session}

[Question]
{question}

[Options]
{answer_options}
\end{promptbox}

\end{document}

%% file: math_commands.tex
\usepackage{amsmath,amsfonts,bm}

\def\eqref#1{equation~\ref{#1}}

\def\1{\bm{1}}

\DeclareMathAlphabet{\mathsfit}{\encodingdefault}{\sfdefault}{m}{sl}
\SetMathAlphabet{\mathsfit}{bold}{\encodingdefault}{\sfdefault}{bx}{n}